\pdfoutput=1
\documentclass[10pt,twocolumn,letterpaper]{article}

\usepackage{iccv}
\usepackage{times}
\usepackage{epsfig}
\usepackage{graphicx}
\usepackage{amsmath}
\usepackage{amssymb}
\usepackage{subcaption}
\usepackage{float}
\graphicspath{ {./images/} }
\usepackage{multirow}
\usepackage[table]{xcolor}
\usepackage{arydshln}
\usepackage{flushend}

\makeatletter
\def\thickhline{%
  \noalign{\ifnum0=`}\fi\hrule \@height \thickarrayrulewidth \futurelet
   \reserved@a\@xthickhline}
\def\@xthickhline{\ifx\reserved@a\thickhline
               \vskip\doublerulesep
               \vskip-\thickarrayrulewidth
             \fi
      \ifnum0=`{\fi}}
\makeatother

\newlength{\thickarrayrulewidth}
\makeatletter
\def\mediumhline{%
  \noalign{\ifnum0=`}\fi\hrule \@height \mediumarrayrulewidth \futurelet
   \reserved@a\@xthickhline}
\def\@xthickhline{\ifx\reserved@a\thickhline
               \vskip\doublerulesep
               \vskip-\thickarrayrulewidth
             \fi
      \ifnum0=`{\fi}}
\makeatother

\newlength{\mediumarrayrulewidth}
\usepackage[pagebackref=true,breaklinks=true,letterpaper=true,colorlinks,bookmarks=false]{hyperref}

\iccvfinalcopy 

\def\iccvPaperID{4217} 

\ificcvfinal\fi
\begin{document}

\title{{\sc AttentionRNN}: A Structured Spatial Attention Mechanism}

\author{Siddhesh Khandelwal\\
University of British Columbia\\
{\tt\small skhandel@cs.ubc.ca}
\and
Leonid Sigal\\
University of British Columbia\\
{\tt\small lsigal@cs.ubc.ca}
}

\maketitle

\begin{abstract}
   Visual attention mechanisms have proven to be integrally important constituent components of many modern deep neural architectures. They provide an efficient and effective way to utilize visual information selectively, which has shown to be especially valuable in multi-modal learning tasks. 
However, all prior attention frameworks lack the ability to explicitly model structural dependencies among attention variables, making it difficult to predict consistent attention masks.
   In this paper we develop a novel {\em structured spatial} attention mechanism which is end-to-end trainable and can be integrated with any feed-forward convolutional neural network. This proposed AttentionRNN layer explicitly enforces structure over the spatial attention variables by sequentially predicting attention values in the spatial mask in a bi-directional raster-scan and inverse raster-scan order. As a result, each attention value depends not only on local image or contextual information, but also on the previously predicted attention values. Our experiments show consistent quantitative and qualitative improvements on a variety of recognition tasks and datasets; including image categorization, question answering and image generation. 
\end{abstract}

\section{Introduction}
In recent years, computer vision has made tremendous progress across many complex recognition 
tasks, including image classification \cite{krizhevsky2012nips,zheng2017learning}, image captioning \cite{chen2017sca,johnson2016cvpr,xu2015show,you2016image}, image generation \cite{tang2014nips,zhang2018arxiv,zhao2018modular} and visual question answering (VQA) \cite{antol2015iccv,fukui2016multimodal,johnson2017cvpr,lu2016hierarchical,seo2017visual,tommasi2014bmvc,xu2016ask,yang2016stacked}. Arguably, much of this success  
can be attributed to the use of visual attention mechanisms which, similar to the human perception, identify the important regions of an image.
Attention mechanisms typically produce a spatial mask for the given image feature tensor. In an ideal scenario, the mask is expected to have higher activation values over the features corresponding to the regions of interest, and lower activation values everywhere else. For tasks that are multi-modal in nature, like VQA, a query (\eg, a question)
can additionally be  used as an input to generate the mask. In such cases, the attention activation is usually a function of similarity between the corresponding encoding of the image region and the question in a pre-defined or learned embedding space. 

\begin{figure}[t]
\centering
  \includegraphics[width=0.8\linewidth]{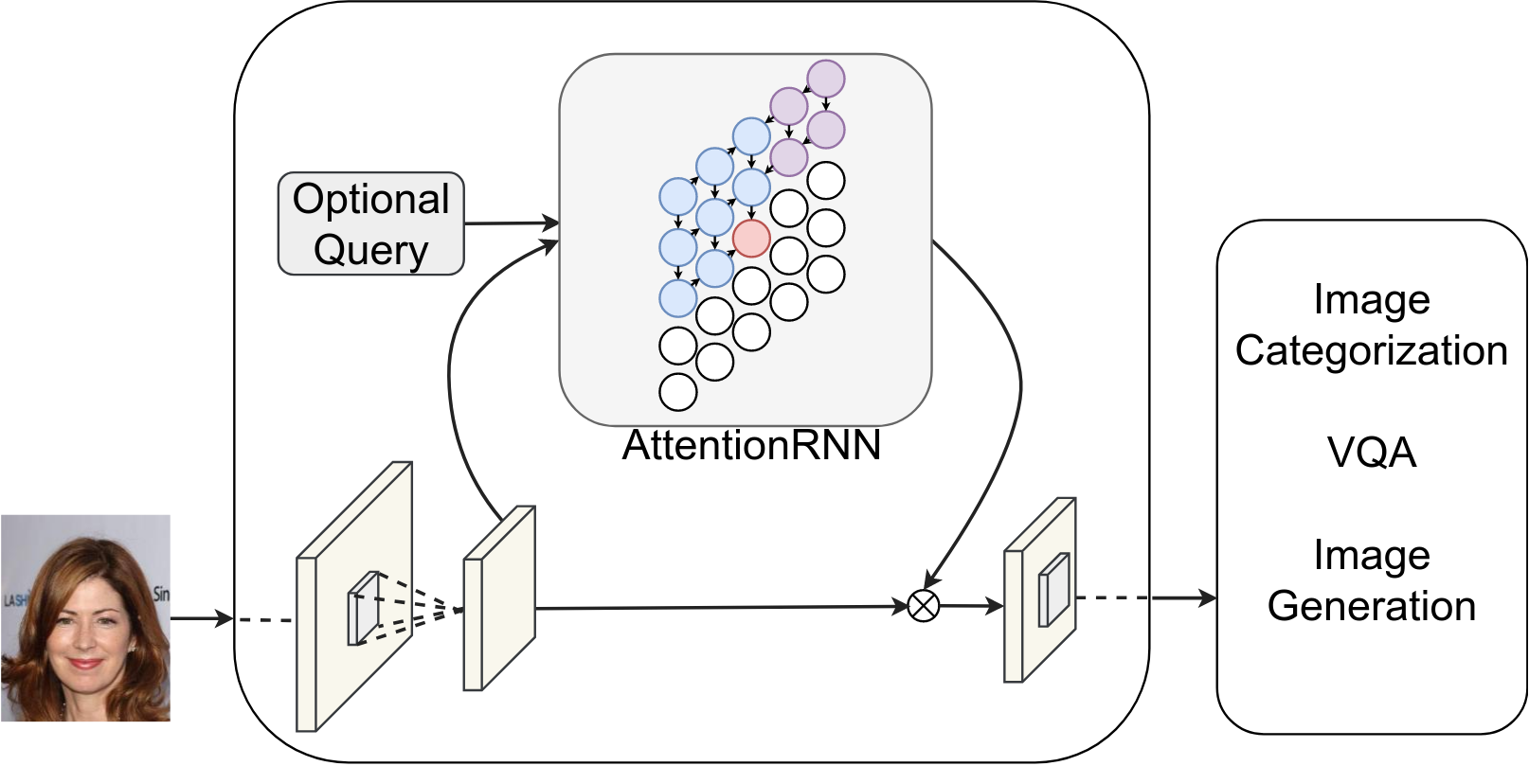}
    \vspace{-0.1in}
  \caption{{\bf AttentionRNN.} Illustration of the proposed structured attention network as a module for down stream task.}
  \label{fig:model}
 \vspace{-0.1in}
\end{figure}
\begin{figure*}[t]
\centering
\begin{subfigure}{.2\textwidth}
  \centering
  \includegraphics[width=\linewidth]{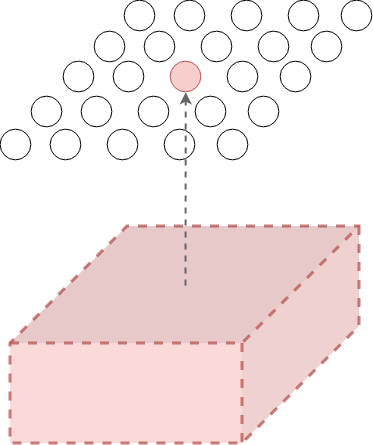}
  \caption{{\em Global} Attention}
  \label{subfig1:att}
\end{subfigure}
\begin{subfigure}{.2\textwidth}
  \centering
  \includegraphics[width=\linewidth]{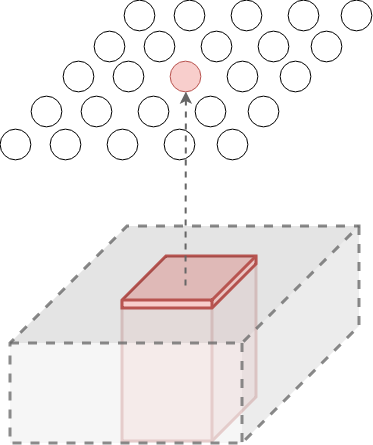}
  \caption{{\em Local} Attention}
  \label{subfig2:att}
\end{subfigure}
\begin{subfigure}{.2\textwidth}
  \centering
  \includegraphics[width=\linewidth]{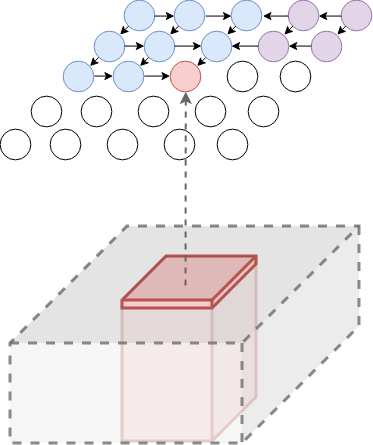}
  \caption{{\em Structured} Attention}
  \label{subfig3:att}
\end{subfigure}
\vspace{-0.15in}
\caption{{\bf Different types of attention mechanisms.} Compared are (a) {\em global} and (b) {\em local} attention mechanisms explored in  prior works and proposed {\em structured} AttentionRNN architecture in (c).}
\label{fig:att}
\vspace{-0.2in}
\end{figure*}

Existing visual attention mechanisms can be broadly characterized into two categories: {\em global} or {\em local}; see Figure~\ref{subfig1:att} and \ref{subfig2:att} respectively for illustration. 
Global mechanisms predict all the attention variables jointly, typically based on a dense representation of the image feature map. Such mechanisms are prone to overfitting and are only computationally feasible for low-resolution image features. Therefore, typically, these are only applied at the last convolutional layer of a CNN \cite{lu2016hierarchical, zhu2016visual7w}. 
The local mechanisms generate attention values for each spatial attention variable independently based on corresponding image region \cite{fukui2016multimodal,seo2017visual,seo2016progressive} (\ie, feature column) or with the help of local context \cite{seo2016progressive, woo2018cbam, zhao2018modular}. As such, local attention mechanisms can be applied at arbitrary resolution and can be used at various places within a CNN network (\eg, in \cite{seo2016progressive} authors use them before each sub-sampling layer and in \cite{woo2018cbam} as part of each residual block).
However, all the aforementioned models lack explicit structure in the generated attention masks. This is often exhibited by lack of coherence or sharp discontinuities in the generated attention activation values \cite{seo2016progressive}. 

Consider a VQA model attending to regions required to answer the question, ``Do the two spheres next to each other have the same color?''. Intuitively, attention mechanisms should focus on the two spheres in question. Furthermore, attention region corresponding to one sphere should inform the estimates for attention region for the other, both in terms of shape and size. However, most traditional attention mechanisms have no ability to encode such dependencies. Recent modularized architectures \cite{andreas2016cvpr,hu2018eccv} are able to address some of these issues with attentive {\em reasoning}, but they are relevant only for a narrow class of VQA problems. Such models are inapplicable to scenarios involving self-attention \cite{woo2018cbam} or generative architectures, where granular shape-coherent attention masks are typically needed \cite{zhao2018modular}.

In this paper, we argue that these challenges can be addressed by {\em structured} spatial attention. Such class of attention models can potentially encode arbitrary constraints between attention variables, be it top-down structured knowledge or local/global consistency and dependencies. To enforce this structure, we propose a novel attention mechanism which we refer to as AttentionRNN (see Figure \ref{subfig3:att} for illustration). We draw inspiration from the Diagonal BiLSTM architecture proposed in \cite{van2016pixel}. As such, AttentionRNN generates a spatial attention mask by traversing the image diagonally, starting from a corner at the top and going to the opposite corner at the bottom. When predicting the attention value for a particular image feature location, structure is enforced by taking into account: (i) local image context around the corresponding image feature location and, 
more importantly, (ii) information about previously generated attention values. 

One of the key benefits of our model is that it can be used agnostically in any existing feed-forward neural network at one or multiple convolutional feature levels (see Figure \ref{fig:model}). To support this claim, we evaluate our method on different tasks and with different backbone architectures.
For VQA, we consider the Progressive Attention Network (PAN) \cite{seo2016progressive} and Multimodal Compact Bilinear Pooling (MCB) \cite{fukui2016multimodal}. For image generation, we consider the Modular Generative Adversarial Networks (MGAN) \cite{zhao2018modular}. For image categorization, we consider the Convolutional Block Attention Module (CBAM) \cite{woo2018cbam}. When we replace the existing attention mechanisms in these models with our proposed AttentionRNN, we observe higher overall performance along with better spatial attention masks.

\vspace{0.02in}
\noindent
{\bf Contributions:} Our contributions can be summarized as follows: (1) We propose a novel spatial attention mechanism which explicitly encodes structure over the spatial attention variables by sequentially predicting values.
As a consequence, each attention value in a spatial mask depends not only on local image or contextual information, but also on previously predicted attention values. (2) We illustrate that this general attention mechanism can work with any existing model that relies on, or can benefit from, spatial attention; showing its effectiveness on a variety of different tasks and datasets. (3) Through experimental evaluation, we observe improved performance and better attention masks on VQA, image generation and image categorization tasks.

\section{Related Work}
\subsection{Visual Attention}
Visual attention mechanisms have been widely adopted in the computer vision community owing to their ability to focus on important regions in an image. Even though there is a large variety of methods that deploy visual attention, they can be categorized based on key properties of the underlying attention mechanisms. For ease of understanding, we segregate related research using these properties.

\vspace{0.02in}
\noindent 
\textbf{Placement of attention in a network}. Visual attention mechanisms are typically applied on features extracted by a  convolutional neural network (CNN). Visual attention can either be applied: (1) at the end of a CNN network, or (2) iteratively at different layers within a CNN network. 

Applying visual attention at the end of a CNN network is the most straightforward way of incorporating visual attention in deep models. This has led to an improvement in model performance across a variety of computer vision tasks, including image captioning \cite{chen2017sca, xu2015show, you2016image}, image recognition \cite{zheng2017learning}, VQA \cite{lu2016hierarchical, xu2016ask, yang2016stacked, zhu2016visual7w}, and visual dialog \cite{seo2017visual}.

On the other hand, there have been several approaches that iteratively apply visual attention, operating over multiple CNN feature layers \cite{jaderberg2015spatial, seo2016progressive, woo2018cbam}. Seo \etal \cite{seo2016progressive} progressively apply attention after each pooling layer of a CNN network to accurately attend over target objects of various scales and shape. Woo \etal \cite{woo2018cbam} use a similar approach, but instead apply two different types of attention - one that attends over feature channels and the other that attends over the spatial domain.

\vspace{0.02in}
\noindent
\textbf{Context used to compute attention}. Attention mechanisms differ on how much information they use to compute the attention mask. They can be \emph{global}, that is use all the available image context to jointly predict the values in an attention mask \cite{lu2016hierarchical, xu2015show}. As an example, \cite{lu2016hierarchical} propose an attention mechanism for VQA where the attention mask is computed by projecting the image features into some latent space and then computing its similarity with the question.

Attention mechanisms can also be \emph{local}, where-in attention for each variable is generated independently or using a corresponding local image region \cite{fukui2016multimodal,seo2017visual,seo2016progressive, woo2018cbam, zhao2018modular}. For example, \cite{seo2016progressive, woo2018cbam, zhao2018modular} use a $k\times k$ convolutional kernel to compute a particular attention value, allowing them to capture local information around the corresponding location.

\vspace{0.02in}
\noindent
None of the aforementioned works enforce structure over the generated attention masks. Structure over the values of an image, however, has been exploited in many autoregressive models trained to generate images. The next section briefly describes the relevant work in this area.

\subsection{Autoregressive Models for Image Generation}
Generative image modelling is a key problem in computer vision. In recent years, there has been significant work in this area \cite{goodfellow2014generative, kingma2014auto, rezende2014stochastic, salimans2017pixelcnn++, van2016pixel, zhang2017stackgan, zhao2018modular}. Although most work uses stochastic latent variable models like VAEs \cite{kingma2014auto, rezende2014stochastic} or GANs \cite{goodfellow2014generative, zhang2017stackgan, zhao2018modular}, autoregressive models \cite{salimans2017pixelcnn++, van2016conditional, van2016pixel} provide a more tractable approach to model the joint distribution over the pixels. These models leverage the inherent structure over the image, which enables them to express the joint distribution as a product of conditional distributions - where the value of the next pixel is dependent on all the previously generated pixels.

Van \etal \cite{van2016pixel} propose a PixelRNN network that uses LSTMs \cite{hochreiter1997long} to model this sequential dependency between the pixels. They also introduce a variant, called PixelCNN, that uses CNNs instead of LSTMs to allow for faster computations. They later extend PixelCNN to allow the model to be conditioned on some query \cite{van2016conditional}. Finally, \cite{salimans2017pixelcnn++} propose further simplifications to the PixelCNN architecture to improve performance. 

Our work draws inspiration from the PixelRNN architecture proposed in \cite{van2016pixel}. We extend PixelRNN to model structural dependencies within attention masks.

\section{Approach}
Given an input image feature $\mathbf{X} \in \mathbb{R}^{h \times m \times n}$, our goal is to predict a spatial attention mask $\mathbf{A} \in \mathbb{R}^{m\times n}$, where $h$ represents the number of channels, and $m$ and $n$ are the number of rows and the columns of $\mathbf{X}$ respectively. Let $\mathbf{X} = \{\mathbf{x}_{1,1}, \dots, \mathbf{x}_{m,n}\}$, where $\mathbf{x}_{i,j} \in \mathbb{R}^h$ be a column feature corresponding to the spatial location $(i,j)$. Similarly, let $\mathbf{A} = \{a_{1,1}, \dots, a_{m,n}\}$, where $a_{i,j} \in \mathbb{R}$ be the attention value corresponding to $\mathbf{x}_{i,j}$. Formally, we want to model the conditional distribution $p(\mathbf{A}~|~\mathbf{X})$. In certain problems, we may also want to condition on other auxiliary information in addition to $\mathbf{X}$, \eg in VQA on a question. 
While in this paper we formulate and model attention probabilistically, most traditional attention models directly predict the attention values, which can be regarded as a point estimate (or expected value) of $\mathbf{A}$ under our formulation. 

Global attention mechanisms \cite{lu2016hierarchical, zhu2016visual7w} predict
$\mathbf{A}$ directly from $\mathbf{X}$
using a fully connected layer. 
Although this makes no assumptions on the factorization of $p(\mathbf{A}~|~\mathbf{X})$, it becomes intractable as the size of $\mathbf{X}$ increases. This is mainly due to the large number of parameters in the fully connected layer.

Local attention mechanisms \cite{seo2017visual, seo2016progressive, woo2018cbam, zhao2018modular}, on the other hand, make strong independence assumptions on the interactions between the attention variables $a_{i,j}$. Particularly, they assume each attention variable $a_{i,j}$ to be independent of other variables given some local spatial context $\delta(\mathbf{x}_{i, j})$. More formally, for local attention mechanisms,
\begin{align}
    \begin{split}
        p\left(\mathbf{A}~|~\mathbf{X}\right) &\approx \prod_{i=1, j=1}^{i=m, j=n} p\left(a_{i,j}~|~ \delta(\mathbf{x}_{i,j})\right)
    \end{split}
\end{align}
Even though such a factorization improves tractability, the strong independence assumption often leads to attention masks that lack coherence and contain sharp discontinuities.

Contrary to local attention mechanisms, our proposed \emph{AttentionRNN} tries to capture some of the structural dependencies between the attention variables $a_{i,j}$. We assume 
\begin{align}
    \label{eq:chainrule}
    p(\mathbf{A}~|~\mathbf{X}) &= \prod_{i=1, j=1}^{i=m, j=n} p\left(a_{i,j} ~|~ \mathbf{a}_{<i,j},~ \mathbf{X})\right)\\
    \label{eq:arrnprob}
    &\approx \prod_{i=1, j=1}^{i=m, j=n} p\left(a_{i,j} ~|~ \mathbf{a}_{<i,j},~ \delta(\mathbf{x}_{i,j})\right)
\end{align}
where $\mathbf{a}_{<i,j} = \{a_{1,1}, \dots, a_{i-1,j}\}$ (The blue and green region in Figure \ref{fig:skewing}). That is, each attention variable $a_{i,j}$ is now dependent on : (i) the local spatial context $\delta(\mathbf{x}_{i,j})$, and (ii) all the previous attention variables $\mathbf{a}_{<i,j}$. Note that Equation \ref{eq:chainrule} is just a direct application of the chain rule. Similar to local attention mechanisms, and to reduce the computation overhead, we assume that a local spatial context $\delta(\mathbf{x}_{i,j})$ is a sufficient proxy for the image features $\mathbf{X}$ when computing $a_{i,j}$. Equation \ref{eq:arrnprob} describes the final factorization we assume. 

One of the key challenges in estimating $\mathbf{A}$ based on Equation \ref{eq:arrnprob} is to efficiently compute the term $\mathbf{a}_{<i,j}$. A straightforward solution is to use a recurrent neural network (\eg LSTMs) to summarize the previously predicted attention values $\mathbf{a}_{<i,j}$ into its hidden state. This is a common approach employed in many sequence prediction methods \cite{bahdanau2014neural, shankar2018posterior, venugopalan2015sequence}. However, while sequences have a well defined ordering, image features can be traversed in multiple ways due to their spatial nature. Naively parsing the image along its rows using an LSTM, though provides an estimate for $\mathbf{a}_{<i,j}$, fails to correctly encode the necessary information required to predict $a_{i,j}$. 
As an example, the LSTM will tend to forget information from the neighbouring variable $a_{i-1,j}$ as it was processed $n$ time steps ago.

To alleviate this issue, \emph{AttentionRNN} instead parses the image in a diagonal fashion, starting from a corner at the top and going to the opposite corner in the bottom. It builds upon the Diagonal BiLSTM layer proposed by \cite{van2016pixel} to efficiently perform this traversal. The next section describes our proposed \emph{AttentionRNN} mechanism in detail. 

\subsection{AttentionRNN}\label{sec:attention}
Our proposed structured attention mechanism builds upon the Diagonal BiLSTM layer proposed by \cite{van2016pixel}. We employ two LSTMs, one going from the top-left to bottom-right corner ($\mathcal{L}^{l}$) and the other from the top-right to the bottom-left corner ($\mathcal{L}^r$).

\begin{figure}[t]
\centering
\includegraphics[width=0.48\textwidth]{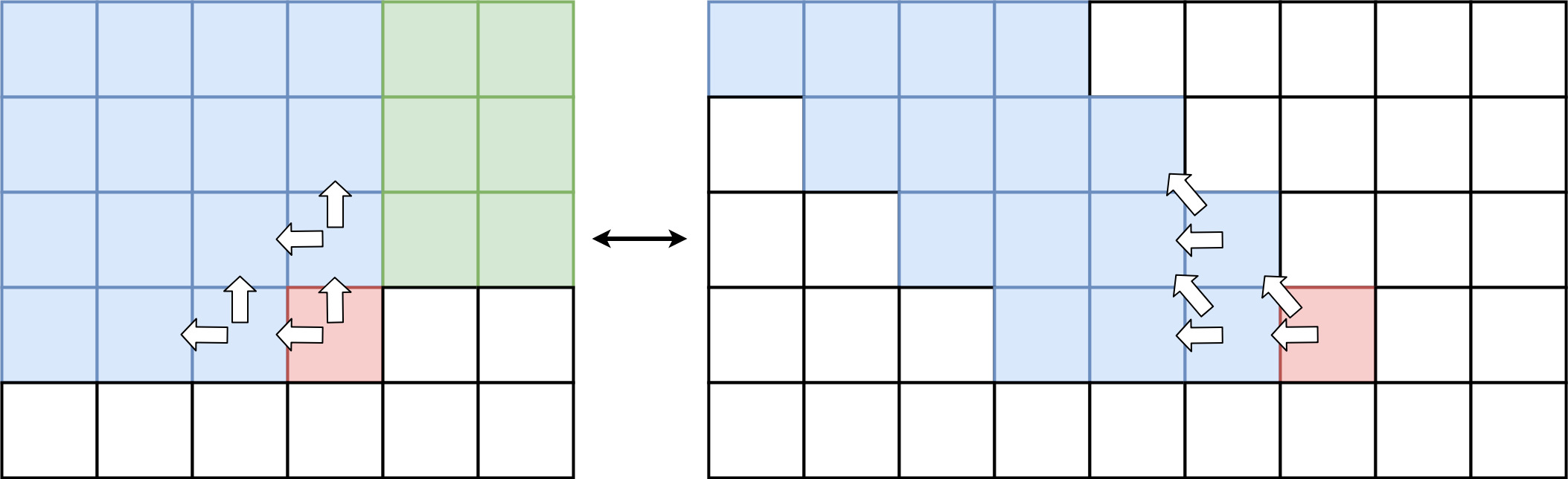}
\caption{{\bf Skewing operation.} This makes it easier to compute convolutions along the diagonal. The arrows indicate dependencies between attention values. To obtain the image on the right, each row of the left image is offset by one position with respect to its previous row.}
\label{fig:skewing}
\vspace{-0.2in}
\end{figure}

As mentioned in Equation \ref{eq:arrnprob}, for each $a_{i,j}$, our objective is to estimate $p\left(a_{i,j} ~|~ \mathbf{a}_{<i,j}, \delta(\mathbf{x}_{i,j})\right)$. We assume that this can be approximated via a combination of two distributions. 
\begin{align}
    \label{eq:decompose}
    \begin{split}
        p\left(a_{i,j}|\mathbf{a}_{<i,j}\right) &= \Gamma\left<p\left(a_{i,j}| \mathbf{a}_{<i,<j}\right),~p\left(a_{i,j}|\mathbf{a}_{<i,>j}\right) \right>
    \end{split}
\end{align}
where $\mathbf{a}_{<i,<j}$ is the set of attention variables to the top and left (blue region in Figure \ref{fig:skewing}) of $a_{i,j}$, $\mathbf{a}_{<i,>j}$ is the set of attention variables to the top and right of $a_{i,j}$ (green region in Figure \ref{fig:skewing}), and $\Gamma$ is some combination function. For brevity, we omit explicitly writing $\delta(\mathbf{x}_{i,j})$. Equation \ref{eq:decompose} is further simplified by assuming that all distributions are Gaussian. 
\begin{align}
    \label{eq:gaussian}
    \begin{split}
        p\left(a_{i,j}|\mathbf{a}_{<i,<j}\right) &\approx \mathcal{N}\left(\mu_{i,j}^l, {\sigma_{i,j}^l} \right)\\
    p\left(a_{i,j}|\mathbf{a}_{<i,>j}\right) &\approx \mathcal{N}\left(\mu_{i,j}^r, {\sigma_{i,j}^r} \right)\\
    p\left(a_{i,j}|\mathbf{a}_{<i,j}\right) &\approx \mathcal{N}\left(\mu_{i,j}, {\sigma_{i,j}} \right)
    \end{split}
\end{align}
where,
\begin{align}
    \label{eq:gaussparams}
    \begin{split}
        (\mu_{i,j}^l, \sigma_{i,j}^l)& = f_l\left(\mathbf{a}_{<i,<j}\right);~~ (\mu_{i,j}^r, \sigma_{i,j}^r) = f_r\left(\mathbf{a}_{<i,>j}\right)\\
        &(\mu_{i,j}, \sigma_{i,j}) = \Gamma\left(\mu_{i,j}^l, \sigma_{i,j}^l, \mu_{i,j}^r, \sigma_{i,j}^r \right)
    \end{split}
\end{align}
$f_l$ and $f_r$ are fully connected layers. Our choice for the combination function $\Gamma$ is explained in Section \ref{sec:combination}. For each $a_{i,j}$, $\mathcal{L}^l$ is trained to estimate $(\mu_{i,j}^l, \sigma_{i,j}^l)$, and $\mathcal{L}^r$ is trained to estimate $(\mu_{i,j}^r, \sigma_{i,j}^r)$. We now explain the computation for $\mathcal{L}^l$. $\mathcal{L}^r$ is analogous and has the same formulation.

$\mathcal{L}^l$ needs to correctly approximate $\mathbf{a}_{<i,<j}$ in order to obtain a good estimate of $(\mu_{i,j}^l, \sigma_{i,j}^l)$. As we are parsing the image diagonally, from Figure \ref{fig:skewing} it can be seen that the following recursive relation holds,
\begin{align}
\label{eq:recursion}
    \mathbf{a}_{<i,<j} = f(\mathbf{a}_{<i-1,<j}~~,~~ \mathbf{a}_{<i,<j-1})
\end{align}
That is, for each location $(i,j)$, $\mathcal{L}^l$ only needs to consider two attention variables- one above and the other to the left; \cite{van2016pixel} show that this is sufficient for it to be able to obtain information from all the previous attention variables.

To make computations along the diagonal easier, similar to \cite{van2016pixel}, we first skew $\mathbf{X}$ into a new image feature $\mathbf{\widehat{X}}$. Figure \ref{fig:skewing} illustrates the skewing procedure. Each row of $\mathbf{X}$ is offset by one position with respect to the previous row. $\mathbf{\widehat{X}}$ is now an image feature of size $h\times m \times (2n - 1)$. Traversing $\mathbf{X}$ in a diagonal fashion from top left to bottom right is now equivalent to traversing $\mathbf{\widehat{X}}$ along its columns from left to right. As spatial locations $(i-1,j)$ and $(i,j-1)$ in $\mathbf{X}$ are now in the same column in $\widehat{\mathbf{X}}$, we can implement the recursion described in Equation \ref{eq:recursion} efficiently by performing computations on an entire column of $\widehat{\mathbf{X}}$ at once.

Let $\mathbf{\widehat{X}}_j$ denote the $j^{th}$ column of $\mathbf{\widehat{X}}$. Also, let $\mathbf{\widehat{h}}_{j-1}^l$ and $\mathbf{\widehat{c}}_{j-1}^l$ respectively denote the hidden and memory state of $\mathcal{L}^l$ before processing $\mathbf{\widehat{X}}_j$. Both $\mathbf{\widehat{h}}_{j-1}^l$ and $\mathbf{\widehat{c}}_{j-1}^l$ are tensors of size $t\times m$, where $t$ is the number of latent features. The new hidden and memory state is computed as follows.
\begin{align} \label{eq:gates}
    \begin{split}
        [\mathbf{o}_j, \mathbf{f}_j, \mathbf{i}_j, \mathbf{g}_j] &= \sigma\left(\mathbf{K}^{h} \circledast \mathbf{\widehat{h}}_{j-1}^l + \mathbf{K}^{x} \circledast \mathbf{\widehat{X}}_{j}^c\right)\\
        \mathbf{\widehat{c}}_{j}^l &= \mathbf{f}_j \odot \mathbf{\widehat{c}}_{j-1}^l + \mathbf{i}_j \odot \mathbf{g}_j\\
        \mathbf{\widehat{h}}_{j}^l &= \mathbf{o}_j \odot \text{tanh}(\mathbf{c}_{j}^l)
    \end{split}
\end{align}
Here $\circledast$ represents the convolution operation and $\odot$ represents element-wise multiplication. $\mathbf{K}^h$ is a $2 \times 1$ convolution kernel which effectively implements the recursive relation described in Equation \ref{eq:recursion}, and $\mathbf{K}^x$ is a $1 \times 1$ convolution kernel. Both $\mathbf{K}^h$ and $\mathbf{K}^u$ produce a tensor of size $4t \times m$. $\mathbf{\widehat{X}}_{j}^c$ is the $j^{th}$ column of the skewed local context $\mathbf{\widehat{X}}^c$, which is obtained as follows.
\begin{align}
\label{eq:localcontext}
    \begin{split}
        \mathbf{\widehat{X}}^c = \text{skew}\left(\mathbf{K}^c \circledast \mathbf{X}\right)
    \end{split}
\end{align}
where $\mathbf{K}^c$ is a convolutional kernel that captures a $\delta$-size context. For tasks that are multi-modal in nature, a query $\mathbf{Q}$ can additionally be used to condition the generation of $a_{i,j}$. This allows the model to generate different attention mask for the same image features depending on $\mathbf{Q}$. For example, in tasks like VQA, the relevant regions of an image will depend on the question asked. The nature of $\mathbf{Q}$ will also dictate the encoding procedure. As an example, if $\mathbf{Q}$ is a natural language question, it can be encoded using a LSTM layer. $\mathbf{Q}$ can be easily incorporated into \emph{AttentionRNN} by concatenating it with $\mathbf{\widehat{X}}^c$ before passing it to Equation \ref{eq:gates}.

Let $\mathbf{\widehat{h}}^l = \{\mathbf{\widehat{h}}_{1}^l,\dots, \mathbf{\widehat{h}}_{2n-1}^l\}$ be the set of all hidden states obtained from $\mathcal{L}^l$, and $\mathbf{h}^l$ be the set obtained by applying the reverse skewing operation on $\mathbf{\widehat{h}}^l$.
For each $a_{i,j}$, $\mathbf{a}_{<i,<j}$ is then simply the $(i,j)$ spatial element of $\mathbf{h}^l$. $\mathbf{a}_{<i,>j}$ can be obtained by repeating the aforementioned process for $\mathcal{L}^r$, which traverses $\mathbf{X}$ from top-right to bottom-left. Note that this is equivalent to running $\mathcal{L}^r$ from top-left to bottom-right after mirroring $\mathbf{X}$ along the column dimension, and then mirroring the output hidden states $\mathbf{h}^r$ again. Similar to \cite{van2016pixel}, $\mathbf{h}^r$ is shifted down by one row to prevent $\mathbf{a}_{<i,>j}$ from incorporating future attention values.

Once $\mathbf{a}_{<i,<j}$ and $\mathbf{a}_{<i,>j}$ are computed (as discussed above), we can obtain the Gaussian distribution for the attention variable $\mathcal{N}\left(\mu_{i,j}, \sigma_{i,j} \right)$ by following Equation \ref{eq:gaussparams}.
The attention $a_{i,j}$ could then be obtained by either sampling a value from $\mathcal{N}\left(\mu_{i,j}, \sigma_{i,j} \right)$ or simply by taking the expectation and setting $a_{i,j} = \mu_{i,j}$. 
For most problems, as we will see in the experiment section, taking the expectation is going to be most efficient and effective. However, sampling maybe useful in cases where attention is inherently multi-modal. Focusing on different modes using coherent masks might be more beneficial in such situations. 

\begin{figure}[t]
\centering
  \includegraphics[width=\linewidth]{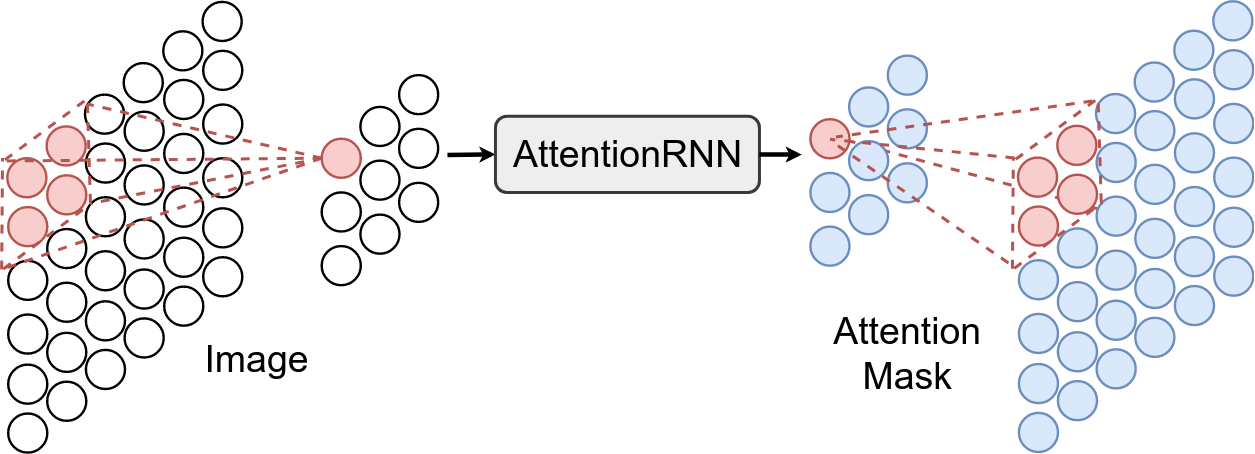}
  \caption{{\bf Block AttentionRNN} for $\gamma=2$. The input is first down-sized using a $\gamma \times \gamma$ convolutional kernel. Attention is computed on this smaller map.}
  \label{fig:upscale}
  \vspace{-0.2in}
\end{figure}
\subsection{Combination Function}\label{sec:combination}
The choice of the combination function $\Gamma$ implicitly imposes some constraints on the interaction between the distributions $\mathcal{N}\left(\mu_{i,j}^l, \sigma_{i,j}^l\right)$ and $\mathcal{N}\left(\mu_{i,j}^r, \sigma_{i,j}^r\right)$. 
For example, assumption of independence would dictate a simple product for $\Gamma$, with resulting operations to produce $(\mu_{i,j}, \sigma_{i,j})$ being expressed in closed form. 
However, it is clear that independence is unlikely to hold due to image correlations. To allow for a more flexible interaction between variables and combination function, 
we instead use a fully connected layer to learn the appropriate $\Gamma$ for a particular task.
\begin{align}
    \label{eq:combination}
    \begin{split}
        \mu_{i,j}, \sigma_{i,j} = f_{comb}\left(\mu_{i,j}^l, \sigma_{i,j}^l, \mu_{i,j}^r, \sigma_{i,j}^r\right)
    \end{split}
\end{align}

\subsection {Block AttentionRNN}\label{sec:upscale}

Due to the poor performance of LSTMs over large sequences, the AttentionRNN layer doesn't scale well to large image feature maps. We introduce a simple modification to the method described in Section \ref{sec:attention} to alleviate this problem, which we refer to as Block AttentionRNN (BRNN). 

BRNN reduces the size of the input feature map $\mathbf{X}$ before computing the attention mask. This is done by splitting $\mathbf{X}$ into smaller blocks, each of size $\gamma \times \gamma$. This is equivalent to down-sampling the original image $\mathbf{X}$ to $\mathbf{X}^{ds}$ as follows.
\begin{align}
    \mathbf{X}^{ds} = \mathbf{K}^{ds} \circledast \mathbf{X}
\end{align}
where $\mathbf{K}^{ds}$ is a convolution kernel of size $\gamma \times \gamma$ applied with stride $\gamma$. In essence, each value in $\mathbf{X}^{ds}$ now corresponds to a $\gamma \times \gamma$ region in $\mathbf{X}$. 

Instead of predicting a different attention probability for each individual spatial location $(i,j)$ in $\mathbf{X}$, BRNN predicts a single probability for each $\gamma \times \gamma$ region. This is done by first computing the attention mask $\mathbf{A}^{ds}$ for the down-sampled image $\mathbf{X}^{ds}$ using AttentionRNN (Section \ref{sec:attention}), and then $\mathbf{A}^{ds}$ is then scaled up using a transposed convolutional layer to obtain the attention mask $\mathbf{A}$ for the original image feature $\mathbf{X}$. Figure \ref{fig:upscale} illustrates the BRNN procedure.

BRNN essentially computes a coarse attention mask for $\mathbf{X}$. Intuitively, this coarse attention can be used in the first few layers of a deep CNN network to identify the key region blocks in the image. The later layers can use this coarse information to generate a more granular attention mask.

\section{Experiments}\label{experiments}
To show the efficacy and generality of our approach, we conduct experiments over four different tasks: visual attribute prediction, image classification, visual question answering (VQA) and image generation. 
We highlight that our goal is not to necessarily obtain the absolute highest raw performance (although we do in many of our experiments), but to show improvements from integrating AttentionRNN into existing state-of-the-art models across a variety of tasks and architectures. Due to space limitations, all model architectures and additional visualizations are described in the supplementary material.

\begin{figure}[t]
\centering
\begin{subfigure}{.15\textwidth}
  \centering
  \includegraphics[width=0.9\linewidth]{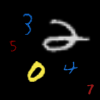}
  \caption{MREF}
  \label{subfig1:synthetic}
\end{subfigure}
\begin{subfigure}{.15\textwidth}
  \centering
  \includegraphics[width=0.9\linewidth]{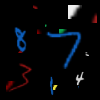}
  \caption{MDIST}
  \label{subfig2:synthetic}
\end{subfigure}
\begin{subfigure}{.15\textwidth}
  \centering
  \includegraphics[width=0.9\linewidth]{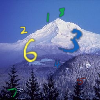}
  \caption{MBG}
  \label{subfig3:synthetic}
\end{subfigure}
\vspace{-0.1in}
\caption{{\bf Synthetic Dataset Samples.} Example images taken from the three synthetic datasets proposed in \cite{seo2017visual}.}
\label{fig:synthetic}
\vspace{-0.1in}
\end{figure}

\subsection{Visual Attribute Prediction}\label{subsec:vap}

\noindent 
\textbf{Datasets.} We experiment on the synthetic MREF, MDIST and MBG datasets proposed in \cite{seo2016progressive}. Figure \ref{fig:synthetic} shows example images from the datasets. The images in the datasets are created from MNIST \cite{lecun1998gradient} by sampling five to nine distinct digits with different colors (green, yellow, white, red, or blue) and varying scales (between 0.5 and 3.0). The datasets have images of size 100 x 100 and only differ in how the background is generated. MREF has a black background, MDIST has a black background with some Gaussian noise, and MBG has real images sampled from the SUN Database \cite{xiao2016sun} as background. The training, validation and test sets contain 30,000, 10,000 and 10,000 images respectively.

\begin{table}[t]
\centering
\begin{tabular}{l|ccc|c}
\thickhline
Attention                  & MREF  & MDIST & MBG   & \begin{tabular}[c]{@{}c@{}}Rel. \\ Runtime\end{tabular} \\ \hline
SAN \cite{xu2015show}                       & 83.42 & 80.06 & 58.07 & 1x                                                      \\
$\lnot \text{CTX}$ \cite{seo2016progressive}        & 95.69 & 89.92 & 69.33 & 1.08x                                                   \\
$\text{CTX}$ \cite{seo2016progressive}               & 98.00 & 95.37 & 79.00 & 1.10x                                                   \\\hdashline
$\text{ARNN}_{ind}^{\sim}$ & 98.72 & 96.70 & 83.68 & \multirow{4}{*}{4.73x}                                  \\
$\text{ARNN}_{ind}$        & 98.58 & 96.29 & 84.27 &                                                         \\
$\text{ARNN}^{\sim}$ & 98.65 & 96.82 & 83.74 &                                                         \\
$\text{ARNN}$        & \textbf{98.93} & \textbf{96.91} & \textbf{85.84} &                                                         \\ 
\thickhline
\end{tabular}
\caption {{\bf Color prediction accuracy.} Results are in \% on MREF, MDIST and MBG datasets. Our AttentionRNN-based model, CNN+ARNN, outperforms all the baselines.}
\label{tab:vapresultspred}
\vspace{-0.2in}
\end{table}

\vspace{0.02in}
\noindent 
\textbf{Experimental Setup.} The performance of AttentionRNN (ARNN) is compared against two {\em local} attention mechanisms proposed in \cite{seo2016progressive}, which are referred as $\lnot  {\text{CTX}}$ and CTX.
ARNN assumes $a_{i,j}=\mu_{i,j}, \delta=3$, where $\mu_{i,j}$ is defined in Equation \ref{eq:combination}.
To compute the attention for a particular spatial location $(i,j)$, CTX uses a $\delta = 3$ local context around $(i,j)$, whereas $\lnot {\text{CTX}}$ only uses the information from location $(i,j)$. We additionally define three variants of ARNN: i) $\text{ARNN}^\sim$ where each $a_{i,j}$ is sampled from $\mathcal{N}\left(\mu_{i,j}, \sigma_{i,j} \right)$, ii) $\text{ARNN}_{ind}$ where the combination function $\Gamma$ assumes the input distributions are independent, and iii) $\text{ARNN}_{ind}^{\sim}$ where $\Gamma$ assumes independence and $a_{i,j}$ is sampled. The soft attention mechanism (SAN) proposed by \cite{xu2015show} is used as an additional baseline. The same base CNN architecture is used for all the attention mechanisms for fair comparison. The CNN is composed of four stacks of $3 \times 3$ convolutions with 32 channels followed by $2 \times 2$ max pooling layer. SAN computes attention only on the output of the last convolution layer, while $\lnot  {\text{CTX}}$, CTX and all variants of ARNN are applied after each pooling layer.  
Given an image, the models are trained to predict the color of the number specified by a query. Chance performance is 20\%.

\begin{figure*}[t]
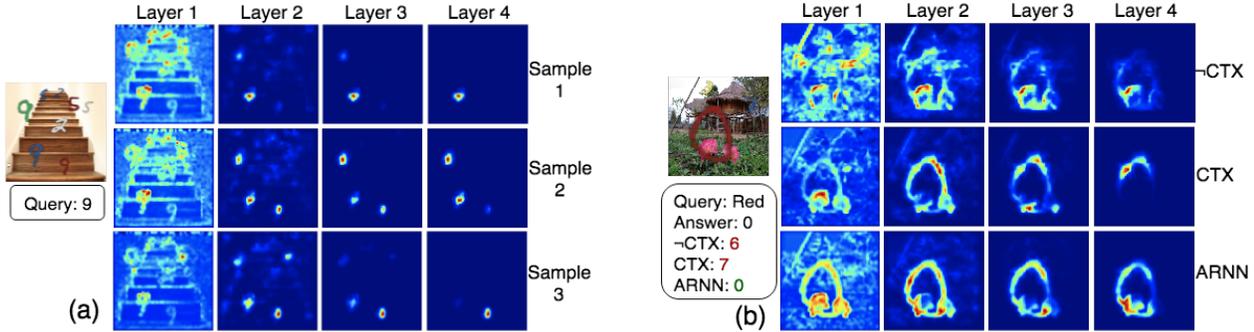

\vspace{-0.1in}
\centering
\begin{subfigure}{0.5\textwidth}
\refstepcounter{subfigure}\label{subfig1:viz}
  \centering
  \includegraphics[width=0.9\linewidth]{vis_vap_1}
\end{subfigure}\begin{subfigure}{0.5\textwidth}
\refstepcounter{subfigure}\label{subfig2:viz}
  \centering
  \includegraphics[width=0.9\linewidth]{vis_inv_vap}
\end{subfigure}

\vspace{-0.1in}
  \caption{{\bf Qualitative analysis of the attention masks.} (a) Layer-wise attended feature maps sampled from $\text{ARNN}^{\sim}$. The samples span all the modes in the image. (b) Layer-wise attended feature maps generated by different mechanisms visualized on images from $\text{MBG}^{inv}$ dataset. Additional visualizations are shown in the supplementary material.}
  \label{fig:qualitative}
  \vspace{-0.2in}
\end{figure*}

\vspace{0.02in}
\noindent 
\textbf{Results.} Table \ref{tab:vapresultspred} shows the color prediction accuracy of various models on MREF, MDIST and MBG datasets. It can be seen that ARNN and all its variants clearly outperform the other baseline methods. The difference in performance is amplified for the more noisy MBG dataset, where ARNN is 6.8\% better than the closest baseline. $\text{ARNN}_{ind}$ performs poorly compared to $\text{ARNN}$, which furthers the reasoning of using a neural network to model $\Gamma$ instead of assuming independence. Similar to \cite{seo2016progressive}, we also evaluate the models on their sensitivity to the size of the target. The test set is divided into five uniform scale intervals for which model accuracy is computed. Table \ref{tab:vapresultsscale} shows the results on the MBG dataset. ARNN is robust to scale variations and performs consistently well on small and large targets. We also test the correctness of the mask generated using the metric proposed by \cite{liu2017attention}, which computes the percentage of attention values in the region of interest. For models that apply attention after each pooling layer, the masks from different layers are combined by upsampling and taking a product over corresponding pixel values. The results are shown for the MBG dataset in Table \ref{tab:vapresultsscale}. ARNN is able to more accurately attend to the correct regions, which is evident from the high correctness score.

\begin{table}[t]
\centering
\setlength\tabcolsep{3.0pt}
\begin{tabular}{@{}l|c|ccccc@{}}
\thickhline
\multirow{2}{*}{Attention} & \multirow{2}{*}{Corr.} & \multicolumn{5}{c}{Scale}                                                          \\
                           &                        & 0.5-1.0        & 1.0-1.5        & 1.5-2.0        & 2.0-2.5        & 2.5-3.0        \\ \hline
SAN \cite{xu2015show}                       & 0.15                   & 53.05          & 74.85          & 72.18          & 59.52          & 54.91          \\
$\lnot\text{CTX}$ \cite{seo2016progressive}          & 0.28                   & 68.20          & 76.37          & 73.30          & 61.60          & 57.28          \\
CTX  \cite{seo2016progressive}                      & 0.31                   & 77.39          & 87.13          & 84.96          & 75.59          & 63.72          \\\hdashline
$\text{ARNN}_{ind}^{\sim}$        & 0.36                   & 82.23          & 89.41          & 86.46          & 84.52          & 81.35          \\
$\text{ARNN}_{ind}$               & 0.34                   & 82.89          & 89.47          & 88.34          & 84.22          & 80.00          \\
$\text{ARNN}^{\sim}$        & 0.39                   & 82.23          & 89.41          & 86.46          & 84.52          & 81.35          \\
$\text{ARNN}$               & \textbf{0.42}          & \textbf{84.45} & \textbf{91.40} & \textbf{86.84} & \textbf{88.39} & \textbf{82.37} \\ \thickhline
\end{tabular}
\caption {{\bf Mask Correctness and Scale experiment on MBG.} The ``Corr.'' column lists the mask correctness metric proposed by \cite{liu2017attention}. The ``Scale'' column shows the color prediction accuracy in \% for different scales.}
\label{tab:vapresultsscale}
\vspace{-0.3in}
\end{table}

From Tables \ref{tab:vapresultspred} and \ref{tab:vapresultsscale}, it can be seen that $\text{ARNN}^{\sim}$ provides no significant advantage over its deterministic counterpart. This can be attributed to the datasets encouraging point estimates, as each input query can only have one correct answer. 
As a consequence, for each $a_{i,j}$, $\sigma_{i,j}$ was observed to underestimate variance. 
However, in situations where an input query can have multiple correct answers, $\text{ARNN}^{\sim}$ can be used to generate diverse attention masks. To corroborate this claim, we test the pre-trained $\text{ARNN}^{\sim}$ on images that are similar to the MBG dataset but have the same digit in multiple colors. Figure \ref{subfig1:viz} shows the individual layer attended feature maps for three different samples from $\text{ARNN}^{\sim}$ for a fixed image and query. For the query ``9'', $\text{ARNN}^{\sim}$ is able to identify the three modes. Note that since $\sigma_{i,j}$'s were underestimated due to aforementioned reasons, they were scaled up before generating the samples. Despite being underestimated $\sigma_{i,j}$'s still capture crucial information. 

\vspace{0.02in}
\noindent
\textbf{Inverse Attribute Prediction.} Figure \ref{subfig1:viz} leads to an interesting observation regarding the nature of the task. Even though $\text{ARNN}^{\sim}$ is able to identify the correct number, it only needs to focus on a tiny part of the target region to be able to accurately classify the color. To further demonstrate the ARNN's ability to model longer dependencies, we test the performance of ARNN, CTX and $\lnot\text{CTX}$ on the $\text{MBG}^{inv}$ dataset, which defines the inverse attribute prediction problem - given a color identify the number corresponding to that color. The base CNN architecture is identical to the one used in the previous experiment. ARNN, CTX and $\lnot\text{CTX}$ achieve an accuracy of 72.77\%, 66.37\% and 40.15\% and a correctness score\cite{liu2017attention} of 0.39, 0.24 and 0.20 respectively. Figure \ref{subfig2:viz} shows layer-wise attended feature maps for the three models. ARNN is able to capture the entire number structure, whereas the other two methods only focus on a part of the target region. Even though CTX uses some local context to compute the attention masks, it fails to identify the complete structure for the number ``0''. A plausible reason for this is that a $3 \times 3$ local context is too small to capture the entire target region. As a consequence, the attention mask is computed in patches. CTX maintains no information about the previously computed attention values, and therefore is unable to assign correlated attention scores for all the different target region patches. ARNN, on the other hand, captures constraints between attention variables, making it much more effective in this situation. 

\begin{table}[t]
\setlength\tabcolsep{2.5pt}
\centering
\begin{tabular}{@{}c|c|ccccc@{}}
\thickhline
                         &                & \multicolumn{5}{c}{Scale}                                                          \\
  & Total          & 0.5-1.0        & 1.0-1.5        & 1.5-2.0        & 2.0-2.5        & 2.5-3.0        \\ \hline
NONE                     & 91.43          & 85.63          & 92.57          & 94.96          & 94.77          & 93.59          \\
ARNN                     & 91.09         &  84.89         & 92.25         & 94.24          &  94.70       & \textbf{94.52}       \\
$\text{BRNN}^{\gamma=3}$ & 91.67          & 85.97          & 93.46          & 94.81          & 94.35          & 93.68         \\
$\text{BRNN}^{\gamma=2}$ & \textbf{92.68} & \textbf{88.10}  & \textbf{94.23}  & \textbf{95.32 } & \textbf{94.80}  & 94.01 \\ \thickhline
\end{tabular}
\caption{{\bf Block AttentionRNN.} Ablation results on $\text{MBG}^\text{b}$ dataset. AttentionRNN (ARNN) and Block AttentionRNN (BRNN) with block sizes of 2 and 3 are compared. }
\label{tab:ablation}
\vspace{-0.2in}
\end{table}

\vspace{0.02in}
\noindent
\textbf{Scalability of ARNN.} The results shown in Table \ref{tab:vapresultspred} correspond to models trained on $100\times100$ input images, where the first attention layer is applied on an image feature of size $50\times50$. To analyze the performance of ARNN on comparatively larger image features, we create a new dataset of $224\times224$ images which we refer to as $\text{MBG}^\text{b}$. The data generation process for $\text{MBG}^\text{b}$ is identical to MBG. We perform an ablation study analyzing the effect of using Block AttentionRNN (BRNN) (Section \ref{sec:upscale}) instead of ARNN on larger image features. For the base architecture, the ARNN model from the previous experiment is 
augmented with an additional stack of convolutional and max pooling layer. The detailed architecture is mentioned in the supplementary material. Table \ref{tab:ablation} shows the color prediction accuracy on different scale intervals for the $\text{MBG}^\text{b}$ dataset. As the first attention layer is now applied on a feature map of size $112\times 112$, ARNN performs worse than the case when no attention (NONE) is applied due to the poor tractability of LSTMs over large sequences. BRNN$^{\gamma=2}$, on the other hand, is able to perform better as it reduces the image feature size before applying attention. However, there is a considerable difference in the performance of BRNN when $\gamma=2$ and $\gamma=3$. When $\gamma=3$, BRNN applies a $3 \times 3$ convolution with stride $3$. This aggressive size reduction causes loss of information.

\subsection{Image Classification}
\vspace{-0.05in}
\noindent 
\textbf{Dataset.} We use the CIFAR-100 \cite{krizhevsky2009learning} to verify the performance of AttentionRNN on the task of image classification. The dataset consists of 60,000 $32\times32$ images from 100 classes. The training/test set contain 50,000/10,000 images. 

\vspace{0.02in}
\noindent 
\textbf{Experimental Setup. }We augment ARNN to the convolution block attention module (CBAM) proposed by \cite{woo2018cbam}. For a given feature map, CBAM computes two different types of attentions: 1) channel attention that exploits the inter-channel dependencies in a feature map, and 2) spatial attention that uses local context to identify relationships in the spatial domain. We replace \emph{only} the spatial attention in CBAM with ARNN. This modified module is referred to as CBAM+ARNN.
ResNet18 \cite{he2016deep} is used as the base model for our experiments. ResNet18+CBAM is the model obtained by using CBAM in the Resnet18 model, as described in \cite{woo2018cbam}. Resnet18+CBAM+ARNN is defined analogously.
We use a local context of $3 \times 3$ to compute the spatial attention for both CBAM and CBAM+ARNN.

\vspace{0.02in}
\noindent 
\textbf{Results. }Top-1 and top-5 error is used to evaluate the performance of the models. The results are summarized in Table \ref{tab:cbam}. CBAM+ARNN provides an improvement of 0.89\% on top-1 error over the closest baseline. Note that this gain, though seems marginal, is larger than what CBAM obtains over ResNet18 with no attention (0.49\% on top-1 error). 
\begin{table}[t]
\setlength\tabcolsep{1.5pt}
\centering
\begin{tabular}{@{}l|cc|c@{}}
\thickhline
                & \begin{tabular}[c]{@{}c@{}}Top-1 \\ Error (\%)\end{tabular} & \begin{tabular}[c]{@{}c@{}}Top-5 \\ Error (\%)\end{tabular} & \begin{tabular}[c]{@{}c@{}}Rel.\\ Runtime\end{tabular} \\ \hline
ResNet18 \cite{he2016deep}        & 25.56                                                         & 6.87                                                         & 1x                                                     \\
ResNet18 + CBAM \cite{woo2018cbam} & 25.07                                                         & 6.57                                                         & 1.43x                                                  \\
ResNet18 + CBAM + ARNN &    \textbf{24.18}                                                            & \textbf{6.42}                                                               & 4.81x                                                   \\ \thickhline
\end{tabular}
\caption {\textbf{Performance on Image Classification.} The Top-1 and Top-5 error \% are shown for all the models. The ARNN based model outperforms all other baselines.}
\label{tab:cbam}
\vspace{-0.1in}
\end{table}

\begin{table}[t]
\centering
\setlength\tabcolsep{2.5pt}
\begin{tabular}{l|cccc|c}
\thickhline
         & Yes/No & Number & Other & Total & \begin{tabular}[c]{@{}c@{}}Rel.\\ Runtime\end{tabular} \\ \hline
MCB \cite{fukui2016multimodal}    & 76.06  & 35.32  & 43.87 & 54.84 & 1x                                                     \\
MCB+ATT \cite{fukui2016multimodal} & 76.12  & 35.84  & 47.84 & 56.89 & 1.66x                                                  \\
MCB+ARNN & \textbf{77.13}  & \textbf{36.75}  & \textbf{48.23} & \textbf{57.58} & 2.46x                                                  \\ \thickhline
\end{tabular}
\caption{\textbf{Performance on VQA.} In \% accuracy.}
\label{tab:vqa}
\vspace{-0.2in}
\end{table}

\subsection{Visual Question Answering}
\noindent
\textbf{Dataset.} We evaluate the performance of ARNN on the task of VQA \cite{antol2015iccv}. The experiments are done on the VQA 2.0 dataset \cite{goyal2017making}, which contains images from MSCOCO \cite{lin2014microsoft} and corresponding questions. As the test set is not publicly available, we evaluate performance 
on the validation set.

\vspace{0.02in}
\noindent 
\textbf{Experimental Setup. }We augment ARNN to the Multimodal Compact Bilinear Pooling (MCB) architecture proposed by \cite{fukui2016multimodal}. This is referred to as MCB+ARNN. Note that even though MCB doesn't give state-of-the-art performance on this task, it is a competitive baseline that allows for easy ARNN integration. MCB+ATT is a variant to MCB that uses a local attention mechanism with $\delta=1$ from \cite{fukui2016multimodal}. 
For fair comparison, MCB+ARNN also uses a $\delta=1$ context.

\vspace{0.02in}
\noindent 
\textbf{Results. } The models are evaluated using the accuracy measure defined in \cite{antol2015iccv}. The results are summarized in Table \ref{tab:vqa}. MCB+ARNN achieves a 0.69\% improvement over the closest baseline. We believe this marginal improvement is because all the models, for each spatial location $(i,j)$, use no context from neighbouring locations (as $\delta=1$).

\subsection{Image Generation}
\noindent
\textbf{Dataset.} We analyze the effect of using ARNN on the task of image generation. Experiments are performed on the CelebA dataset \cite{liu2015faceattributes}, which contains 202,599 face images of celebrities, with 40 binary attributes. The data pre-processing is identical to \cite{zhao2018modular}. The models are evaluated on three attributes: hair color = {\emph{\{black, blond, brown\}}}, gender = {\emph{\{male, female}\}}, and smile = {\emph{\{smile, nosmile\}}}.

\vspace{0.02in}
\noindent
\textbf{Experimental Setup.} We compare ARNN to a local attention mechanism used in the ModularGAN (MGAN) framework \cite{zhao2018modular}. MGAN uses a $3 \times 3$ local context to obtain attention values. We define MGAN+ARNN as the network obtained by replacing the local attention with ARNN. The models are trained to transform an image given an attribute.

\begin{figure}[t]
\centering
\begin{subfigure}{0.45\textwidth}
  \centering
  \includegraphics[width=\linewidth]{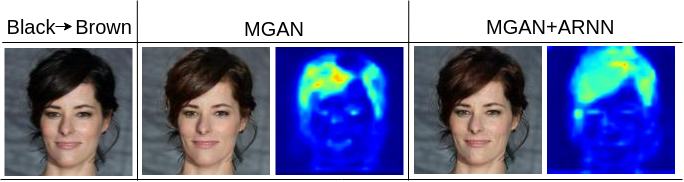}
\end{subfigure}
  \caption{{\bf Qualitative Results on ModularGAN.} Attention masks generated by original ModularGAN \cite{zhao2018modular} and ModularGAN augmented with ARNN are shown. Notice that the hair mask is more uniform for MGAN+ARNN as it is able to encode structural dependencies in the attention mask. Additional results shown in supplementary material.}
  \label{fig:gan}
  \vspace{-0.1in}
\end{figure}

\vspace{0.02in}
\noindent
\textbf{Results.} To evaluate the performance of the models, similar to \cite{zhao2018modular}, we train a ResNet18 \cite{he2016deep} model that classifies the hair color, facial expression and gender on the CelebA dataset. The trained classifier achieves an accuracy of 93.9\%, 99.0\% and 93.7\% on hair color, gender and smile respectively. For each transformation, we pass the generated images through this classifier and compute the classification error (shown in Table \ref{tab:modulargan}). MGAN+ARNN outperforms the baseline on all categories except \emph{hair color}. To analyse this further, we look at the attention masks generated for the \emph{hair color} transformation by both models. As shown in Figure \ref{fig:gan}, we observe that the attention masks generated by MGAN lack coherence over the target region due to discontinuities. MGAN+ARNN, though has a slightly higher classification error, generates uniform activation values over the target region by encoding structural dependencies.

\begin{table}[t]
\vspace{-0.1in}
\centering
\setlength\tabcolsep{6.5pt}
\begin{tabular}{l|ccc|c}
\thickhline
\multicolumn{1}{c|}{} & Hair & Gender & Smile &  \begin{tabular}[c]{@{}c@{}}Rel.\\ Runtime\end{tabular} \\ \hline
MGAN \cite{zhao2018modular}                        & \textbf{2.5}  & 3.2  &  12.6      &1x                                                        \\
MGAN+ARNN                       & 3.0  & \textbf{1.4}  & \textbf{11.4}     &1.96x                                                        \\ \thickhline
\end{tabular}
\caption{\textbf{Performance on Image Generation.} ResNet18 \cite{he2016deep} Classification Errors (\%) for each attribute transformation.
ARNN achieves better performance on two tasks.}
\label{tab:modulargan}
\vspace{-0.2in}
\end{table}

\section{Conclusion}
\vspace{-0.05in}
In this paper, we developed a novel {\em structured} spatial attention mechanism which is end-to-end trainable and can be integrated with any feed-forward convolutional neural network. The proposed AttentionRNN layer explicitly enforces structure over the spatial attention variables by sequentially predicting attention values in the spatial mask. 
Experiments show consistent quantitative and qualitative improvements on a large variety of recognition tasks, datasets and backbone architectures. 

\raggedbottom
\onecolumn
{
    \vspace*{1.0cm}
    \centering
    \Large\bfseries
    Supplementary Material\\
    \vspace*{2.0cm}
}
\setcounter{section}{0}
Section 1 explains the architectures for the models used in the experiments (Section 4 in the main paper). Section 2 provides additional visualizations for the task of Visual Attribute Prediction (Section 4.1 in the main paper) and Image Generation (Section 4.4 in the main paper). These further show the effectiveness of our proposed structured attention mechanism. 

\section{Model Architectures}

\subsection{Visual Attribute Prediction}
Please refer to Section 4.1 of the main paper for the task definition. Similar to \cite{seo2016progressive}, the base CNN architecture is composed of four stacks of $3 \times 3$ convolutions with 32 channels followed by $2 \times 2$ max pooling layer. SAN computes attention only on the output of the last convolution layer, while $\lnot  {\text{CTX}}$, CTX and all variants of ARNN are applied after each pooling layer. Table \ref{tab:modelarc} illustrates the model architectures for each network. \{$\lnot\text{CTX}$, CTX, ARNN\}$_{sigmoid}$ refers to using sigmoid non-linearity on the generated attention mask before applying it to the image features. Similarly, \{$\lnot\text{CTX}$, CTX, ARNN\}$_{softmax}$ refers to using softmax non-linearity on the generated attention mask. We use the same hyper-parameters and training procedure for all models, which is identical to \cite{seo2016progressive}.

For the scalability experiment described in Section 4.1, we add an additional stack of $3 \times 3$ convolution layer followed by a $2 \times 2$ max pooling layer to the ARNN architecture described in Table \ref{tab:modelarc}. This is used as the base architecture. Table \ref{tab:ablationsupp} illustrates the differences between the models used to obtain results mentioned in Table 3 of the main paper.

\begin{table}[h]
\vspace{-0.04in}
\centering
\setlength{\tabcolsep}{2.0em}
\def\arraystretch{1.4}
\begin{tabular}{|c|c|c|c|}
\hline
\textbf{SAN} & \textbf{$\lnot \text{CTX}$} & \textbf{CTX} & \textbf{ARNN} \\ \hline
\multicolumn{4}{|c|}{conv1 (3x3@32)}  \\ \hline
\multicolumn{4}{|c|}{pool1 (2x2)}     \\ \hline
$\downarrow$  & \cellcolor{gray!20}$\lnot \text{CTX}_{sigmoid}$ & \cellcolor{gray!20}CTX$_{sigmoid}$ & \cellcolor{gray!20}ARNN$_{sigmoid}$ \\ \hline
\multicolumn{4}{|c|}{conv2 (3x3@32)}  \\ \hline
\multicolumn{4}{|c|}{pool2 (2x2)}     \\ \hline
$\downarrow$   & \cellcolor{gray!20}$\lnot \text{CTX}_{sigmoid}$ & \cellcolor{gray!20}CTX$_{sigmoid}$ & \cellcolor{gray!20}ARNN$_{sigmoid}$ \\ \hline
\multicolumn{4}{|c|}{conv3 (3x3@32)}  \\ \hline
\multicolumn{4}{|c|}{pool3 (2x2)}     \\ \hline
$\downarrow$   & \cellcolor{gray!20}$\lnot \text{CTX}_{sigmoid}$ & \cellcolor{gray!20}CTX$_{sigmoid}$ & \cellcolor{gray!20}ARNN$_{sigmoid}$ \\ \hline
\multicolumn{4}{|c|}{conv4 (3x3@32)}  \\ \hline
\multicolumn{4}{|c|}{pool4 (2x2)}     \\ \hline
\cellcolor{gray!20}SAN   & \cellcolor{gray!20}$\lnot \text{CTX}_{softmax}$ & \cellcolor{gray!20}CTX$_{softmax}$ & \cellcolor{gray!20}ARNN$_{softmax}$ \\ \hline
\end{tabular}
\vspace{1em}
\vspace{-0.2in}
\caption{Architectures for the models used in Section 4.1 of the main paper. $\downarrow$ implies that the previous and the next layer are directly connected. The input is passed to the top-most layer. The computation proceeds from top to bottom.}
\label{tab:modelarc}
\end{table}

\begin{table}[h]
\centering
\setlength{\tabcolsep}{2.0em}
\def\arraystretch{1.4}
\begin{tabular}{|c|c|c|c|}
\hline
\textbf{NONE} & \textbf{ARNN} & \textbf{BRNN} \\ \hline
\multicolumn{3}{|c|}{conv1 (3x3@32)}  \\ \hline
\multicolumn{3}{|c|}{pool1 (2x2)}     \\ \hline
$\downarrow$  & \cellcolor{gray!20}$\text{ARNN}_{sigmoid}$ & \cellcolor{gray!20}$\text{BRNN}^{\delta}_{sigmoid}$ \\ \hline
\multicolumn{3}{|c|}{\textbf{ARNN} (described in Table \ref{tab:modelarc})}  \\ \hline
\end{tabular}
\vspace{1em}
\caption{Model architectures for the scalability study described in Section 4.1 of the main paper. $\downarrow$ implies that the previous and the next layer are directly connected. \textbf{ARRN} in defined in Table \ref{tab:modelarc}.}
\label{tab:ablationsupp}
\end{table}

\subsection{Image Classification}
\begin{figure}[H]
\centering
\begin{subfigure}{0.85\textwidth}
  \centering
  \includegraphics[width=\linewidth]{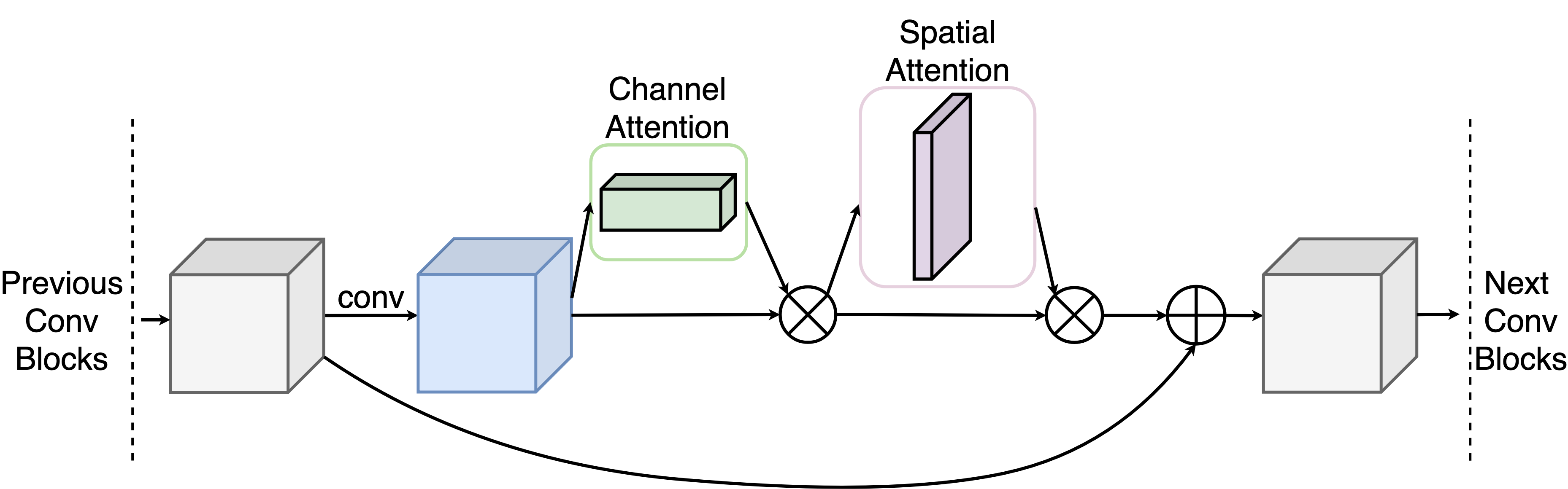}
  \caption{CBAM module}
  \label{subfig1:cbam}
\end{subfigure}
\begin{subfigure}{0.85\textwidth}
  \centering
  \includegraphics[width=\linewidth]{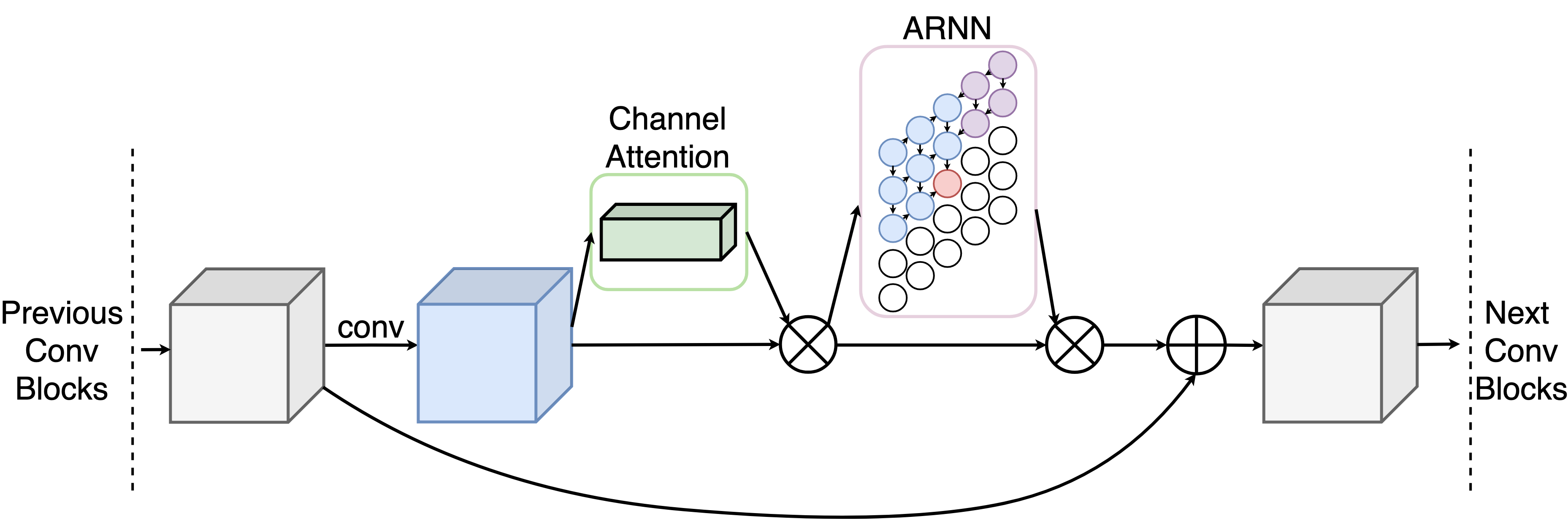}
  \caption{CBAM+ARNN module}
  \label{subfig2:cbam}
\end{subfigure}
\vspace{-0.15in}
\caption{{\bf Difference between CBAM and CBAM+ARNN.} (a) CBAM\cite{woo2018cbam} module integrated with a ResNet\cite{he2016deep} block. (b) CBAM+ARNN replaces the spatial attention in CBAM with ARNN. It is applied similar to (a) after each ResNet\cite{he2016deep} block. Refer to Section 4.2 of the main paper for more details.}
\label{fig:cbam}
\end{figure}

Please refer to Section 4.2 of the main paper for the task definition. We augment ARNN to the convolution block attention module (CBAM) proposed by \cite{woo2018cbam}. For  a  given  feature  map,  CBAM  computes  two  different types  of  attentions:  1)  channel  attention  that  exploits  the inter channel dependencies in a feature map, and 2) spatial attention that uses local context to identify relationships in the spatial domain. Figure \ref{subfig1:cbam} shows the CBAM module integrated with a ResNet \cite{he2016deep} block. We replace only the \emph{spatial attention} in CBAM  with  ARNN.  This  modified  module  is  referred  to as CBAM+ARNN. Figure \ref{subfig2:cbam} better illustrates this modification. Both CBAM and CBAM+ARNN use a local context of $3 \times 3$ to compute attention. We use the same hyper-parameters and training procedure for both CBAM and CBAM+ARNN, which is identical to \cite{woo2018cbam}.

\subsection{Visual Question Answering}
\begin{figure}[H]
\centering
\begin{subfigure}{\textwidth}
  \centering
  \includegraphics[width=\linewidth]{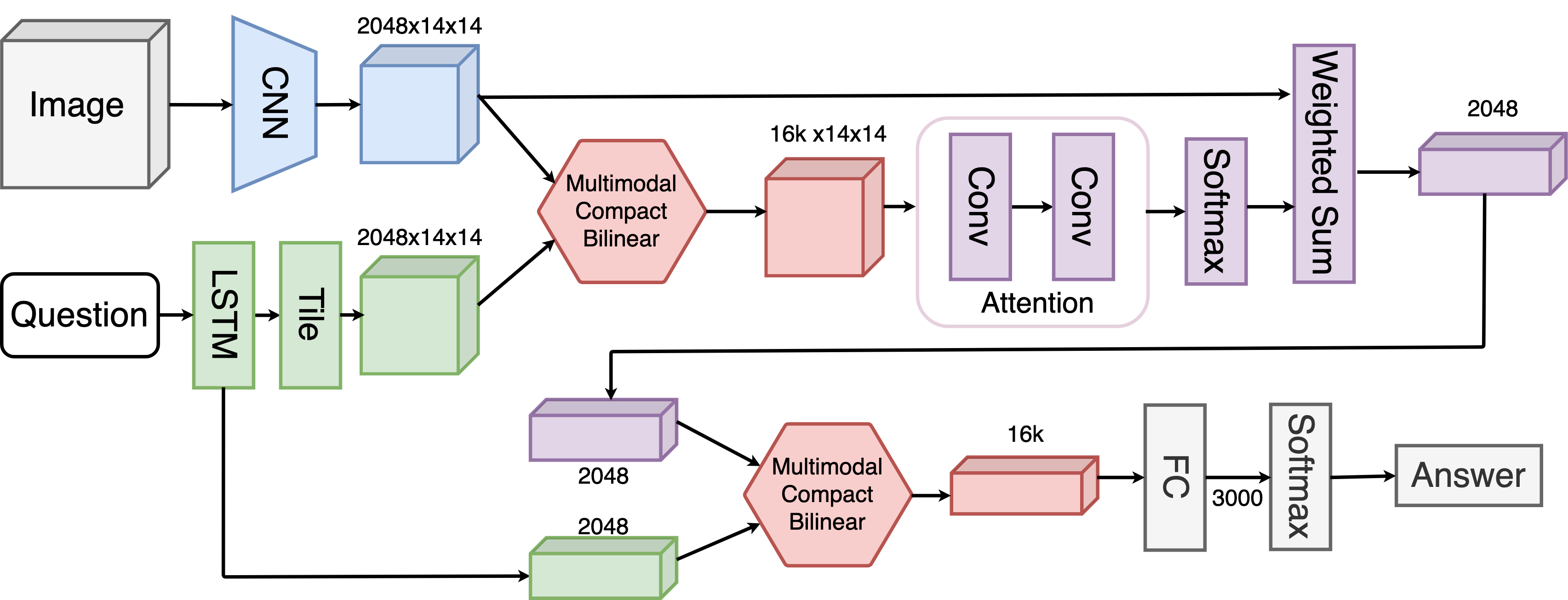}
  \caption{MCB+ATT}
  \label{subfig1:mcb}
\end{subfigure}
\begin{subfigure}{\textwidth}
  \centering
  \includegraphics[width=\linewidth]{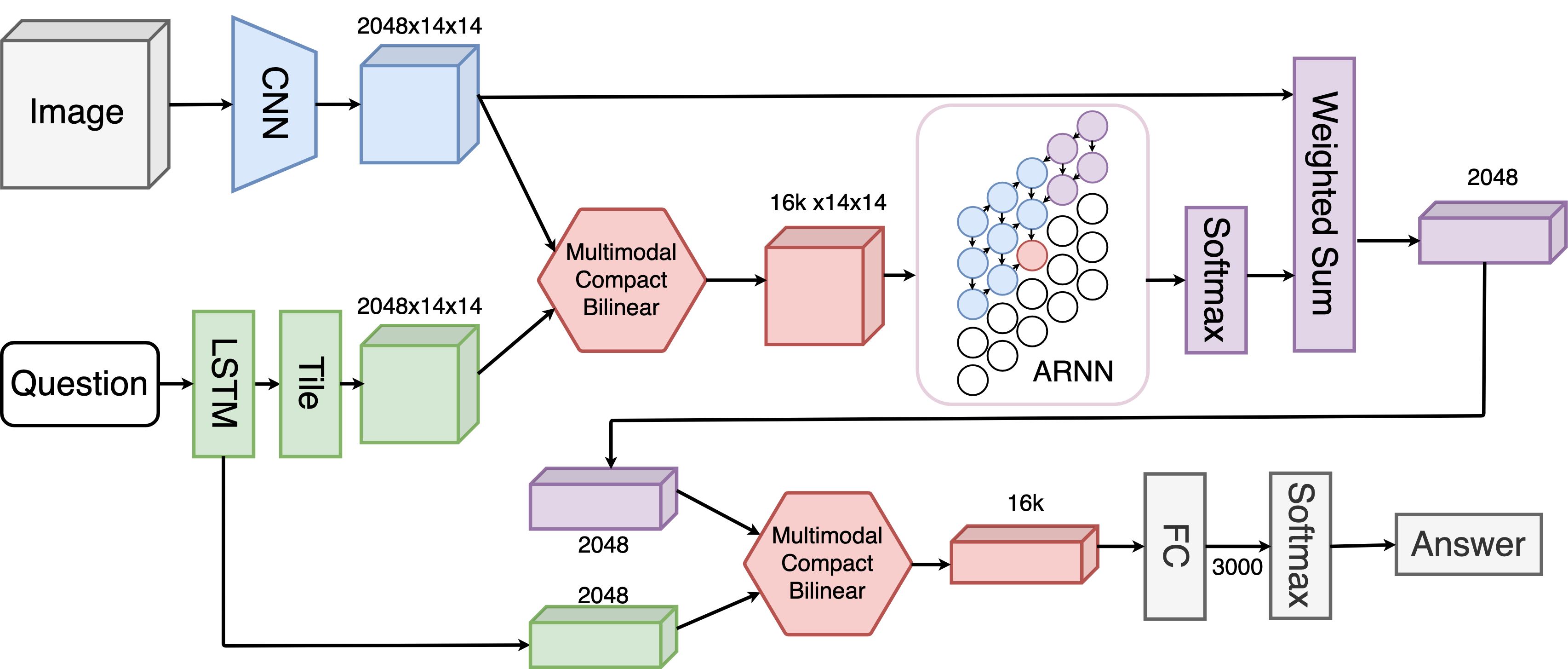}
  \caption{MCB+ARNN}
  \label{subfig2:mcb}
\end{subfigure}
\caption{{\bf Difference between MCB+ATT and MCB+ARNN.} (a) MCB+ATT model architecture proposed by \cite{fukui2016multimodal}. It uses a $1\times 1$ context to compute attention over the image features. (b) MCB+ARNN replaces the attention mechanism in MCB+ATT with ARNN. It is applied in the same location as (a) with $1 \times 1$ context. Refer to Section 4.3 of the main paper for more details.}
\label{fig:mcb}
\end{figure}

Please refer to Section 4.3 of the main paper for task definition. We use the Multimodal Compact Bilinear Pooling with Attention (MCB+ATT) architecture proposed by \cite{fukui2016multimodal} as a baseline for our experiment. To compute attention, MCB+ATT uses two $1 \times 1$ convolutions over the features obtained after using the compact bilinear pooling operation. Figure \ref{subfig1:mcb} illustrates the architecture for MCB+ATT. We replace this attention with ARNN to obtain MCB+ARNN. MCB+ARNN also uses a $1 \times 1$ local context to compute attention. Figure \ref{subfig2:mcb} better illustrates this modification. We use the same hyper-parameters and training procedure for MCB, MCB+ATT and MCB+ARNN, which is identical to \cite{fukui2016multimodal}.

\subsection{Image Generation}
\begin{figure}[H]
\centering
\begin{subfigure}{0.85\textwidth}
  \centering
  \includegraphics[width=\linewidth]{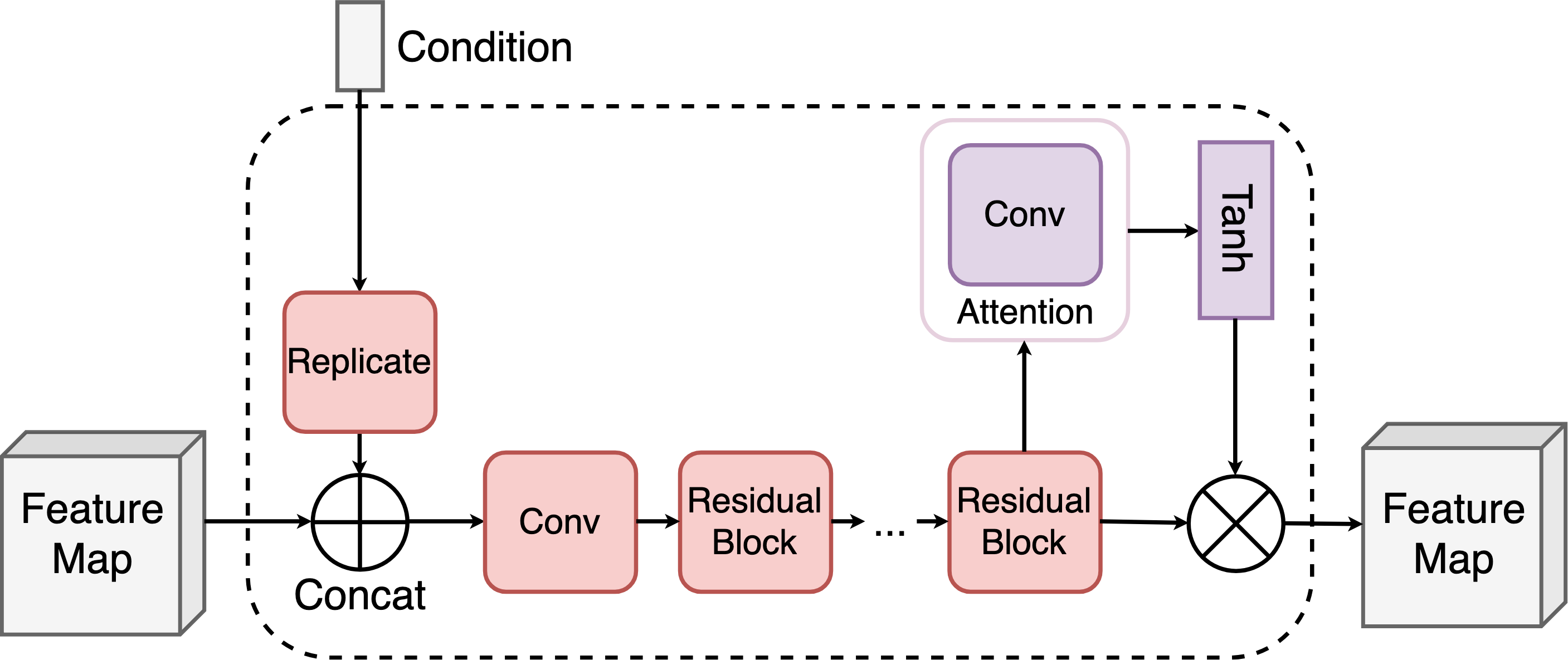}
  \caption{Transformer module for MGAN}
  \label{subfig1:mgan}
\end{subfigure}
\begin{subfigure}{0.85\textwidth}
  \centering
  \includegraphics[width=\linewidth]{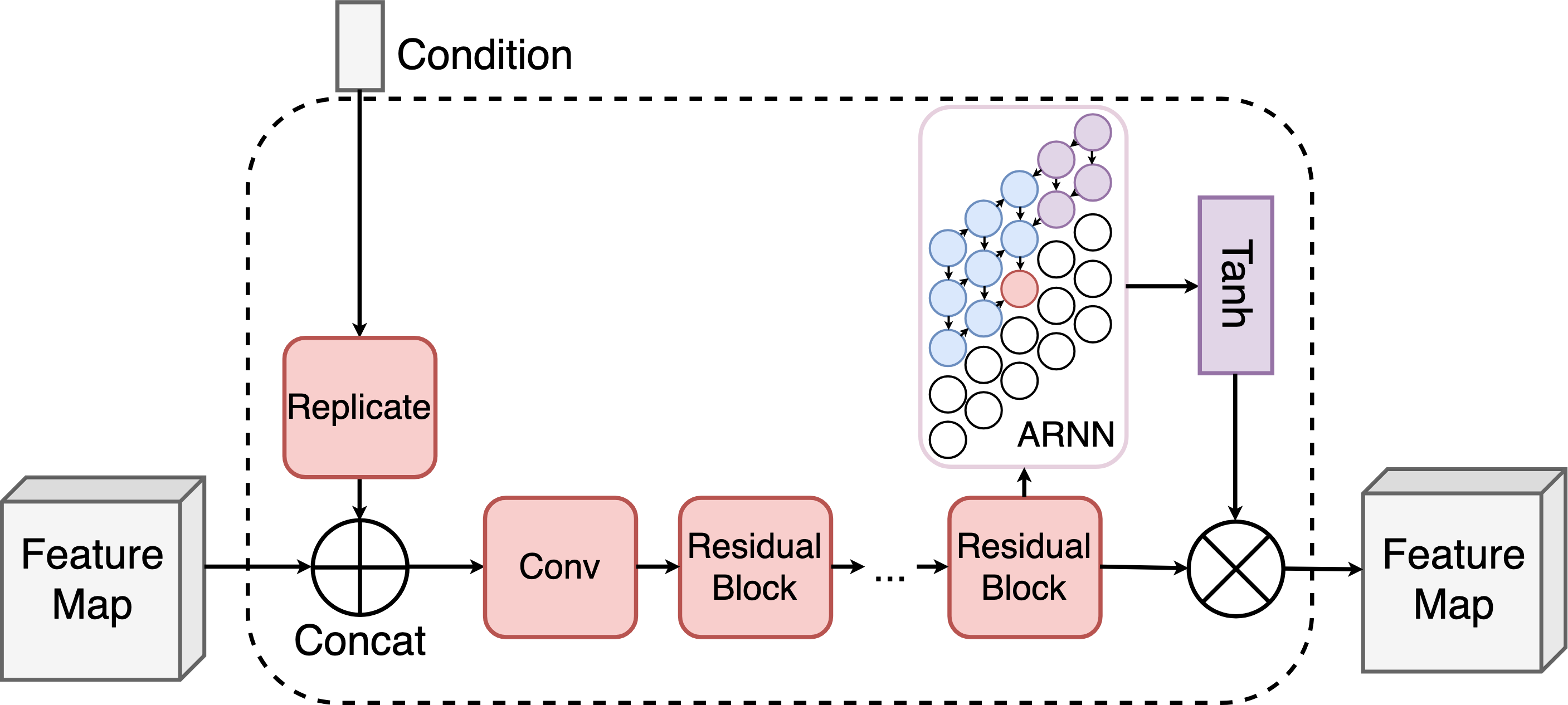}
  \caption{Transformer module for MGAN+ARNN}
  \label{subfig2:mgan}
\end{subfigure}
\caption{{\bf Difference between MGAN and MGAN+ARNN.} (a) The transformer module for the ModularGAN (MGAN) architecture proposed by \cite{zhao2018modular}. It uses a $3\times 3$ local context to compute attention over the intermediate features. (b) MGAN+ARNN replaces the attention mechanism in MGAN with ARNN. It is applied in the same location as (a) with $3 \times 3$ local context. Note that the generator and encoder modules in MGAN and MGAN+ARNN are identical. Refer to Section 4.4 of the main paper for more details.}
\label{fig:mgan}
\end{figure}

Please refer to Section 4.4 of the main paper for task definitions. We compare ARNN to a local attention mechanism used in the ModularGAN (MGAN) framework \cite{zhao2018modular}. MGAN consists of three modules: 1) encoder module that encodes an input image into an intermediate feature representation, 2) generator module that generates an image given an intermediate feature representation as input, and 3) transformer module that transforms a given intermediate representation to a new intermediate representation according to some input condition. The transformer module uses a $3×3$ local context to compute attention over the feature representations. Figure \ref{subfig1:mgan} illustrates the transformer module proposed by \cite{zhao2018modular}. We define MGAN+ARNN as the network obtained by replacing this local attention mechanism in the transformer module with ARNN. Note that the generator and encoder modules are unchanged. MGAN+ARNN also uses a $3 \times 3$ local context to compute attention. Figure \ref{subfig2:mgan} better illustrates this modification to the transformer module. We use the same hyper-parameters and training procedure for both MGAN and MGAN+ARNN, which is identical to \cite{zhao2018modular}.

\newpage
\section{Additional Visualizations}
\subsection{Visual Attribute Prediction}
\begin{figure}[H]
\centering
\includegraphics[width=0.7\linewidth]{sample_7}
\includegraphics[width=0.7\linewidth]{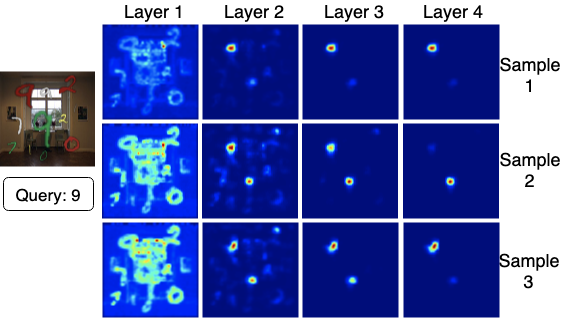}
\caption{{\bf Qualitative Analysis of Attention Masks sampled from $\text{ARNN}^{\sim}$.} Layer-wise attended feature maps sampled from $\text{ARNN}^{\sim}$ for a fixed image and query. The masks are able to span the different modes in the image. For detailed explanation see Section 4.1 of the main paper.}
\label{fig:sample0}
\end{figure}

\begin{figure}[H]
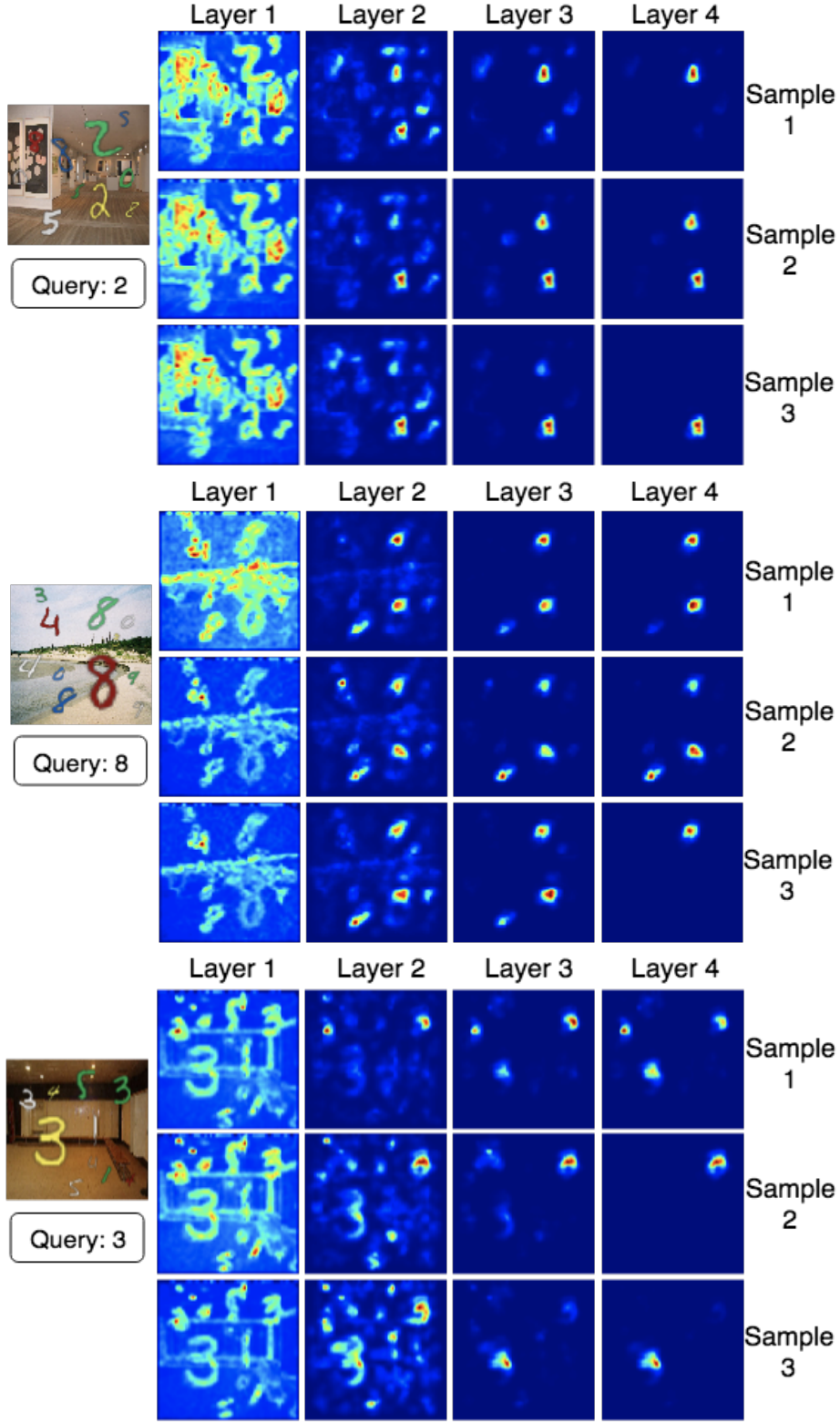

\centering
\includegraphics[width=0.7\linewidth]{sample_4}
\includegraphics[width=0.7\linewidth]{sample_5}
\includegraphics[width=0.7\linewidth]{sample_6}
\caption{{\bf Qualitative Analysis of Attention Masks sampled from $\text{ARNN}^{\sim}$.} Layer-wise attended feature maps sampled from $\text{ARNN}^{\sim}$ for a fixed image and query. The masks are able to span the different modes in the image. For detailed explanation see Section 4.1 of the main paper.}
\label{fig:sample2}
\end{figure}

Please refer to Section 4.1 of the main paper for task definition. Figures \ref{fig:sample0} - \ref{fig:sample2} show the individual layer attended feature maps for three different samples from $\text{ARNN}^{\sim}$ for a fixed image and query. It can be seen that $\text{ARNN}^{\sim}$ is able to identify the different modes in each of the images. 

\begin{figure}[H]
\centering
\includegraphics[width=0.7\linewidth]{sample_1}
\includegraphics[width=0.7\linewidth]{sample_2}
\includegraphics[width=0.7\linewidth]{sample_3}
\caption{{\bf Qualitative Analysis of Attention Masks sampled from $\text{ARNN}^{\sim}$.} Layer-wise attended feature maps sampled from $\text{ARNN}^{\sim}$ for a fixed image and query. The masks are able to span the different modes in the image. For detailed explanation see Section 4.1 of the main paper.}
\label{fig:sample1}
\end{figure}

\subsection{Inverse Attribute Prediction}
\begin{figure}[H]
\centering
\includegraphics[width=0.7\linewidth]{vap_1}
\includegraphics[width=0.7\linewidth]{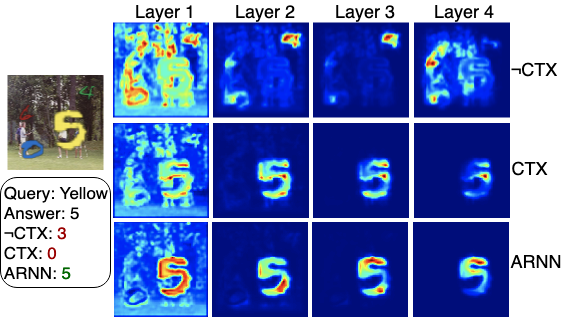}
\caption{{\bf Qualitative Analysis of Attention Masks on $\text{MBG}^{inv}$.} Layer-wise attended feature maps generated by different mechanisms visualized on images from $\text{MBG}^{inv}$ dataset. ARNN is able to capture the entire number structure, whereas the other two methods only focus on a part of the target region or on some background region with the same color as the target number. For detailed explanation see Section 4.1 of the main paper.}
\label{fig:vap0}
\end{figure}

\begin{figure}[H]
\centering
\includegraphics[width=0.7\linewidth]{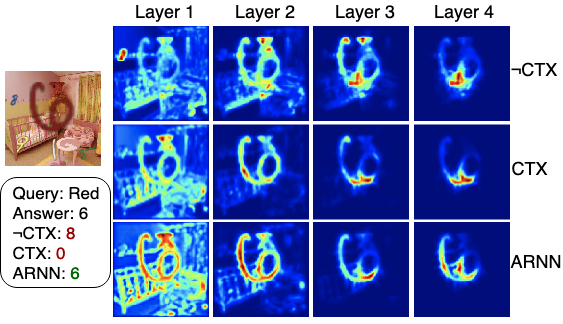}
\includegraphics[width=0.7\linewidth]{vap_2}
\includegraphics[width=0.7\linewidth]{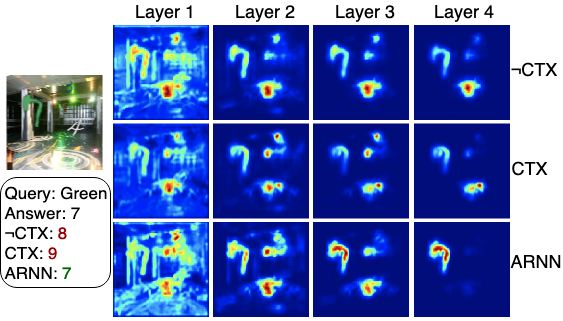}
\caption{{\bf Qualitative Analysis of Attention Masks on $\text{MBG}^{inv}$.} Layer-wise attended feature maps generated by different mechanisms visualized on images from $\text{MBG}^{inv}$ dataset. ARNN is able to capture the entire number structure, whereas the other two methods only focus on a part of the target region or on some background region with the same color as the target number. For detailed explanation see Section 4.1 of the main paper.}
\label{fig:vap1}
\end{figure}

\begin{figure}[H]
\centering
\includegraphics[width=0.7\linewidth]{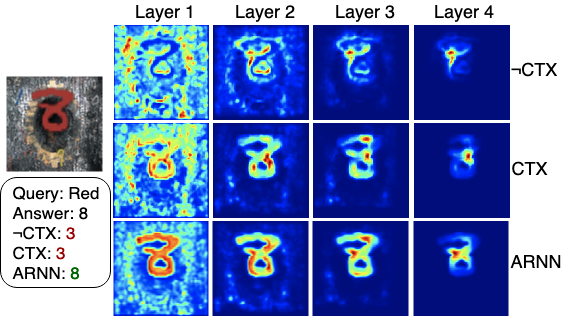}
\includegraphics[width=0.7\linewidth]{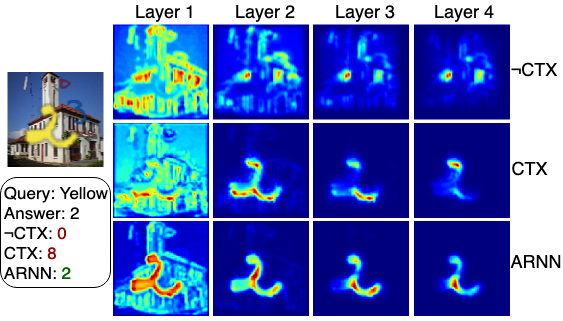}
\includegraphics[width=0.7\linewidth]{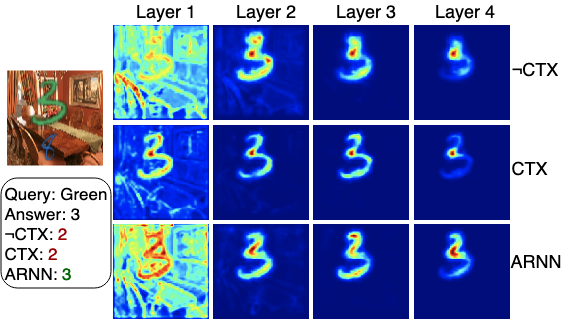}
\caption{{\bf Qualitative Analysis of Attention Masks on $\text{MBG}^{inv}$.} Layer-wise attended feature maps generated by different mechanisms visualized on images from $\text{MBG}^{inv}$ dataset. ARNN is able to capture the entire number structure, whereas the other two methods only focus on a part of the target region or on some background region with the same color as the target number. For detailed explanation see Section 4.1 of the main paper.}
\label{fig:vap2}
\end{figure}

Please refer to Section 4.1 of the main paper for task definition. Figures \ref{fig:vap0} - \ref{fig:vap2} show the individual layer attended feature maps comparing the different attention mechanisms on the $\text{MBG}^{inv}$ dataset. It can be seen that ARNN captures the entire number structure, whereas the other two methods only focus on a part of the target region or on some background region with the same color as the number, leading to incorrect predictions. 

\subsection{Image Generation}
\begin{figure}[H]
\centering
\includegraphics[width=0.8\linewidth]{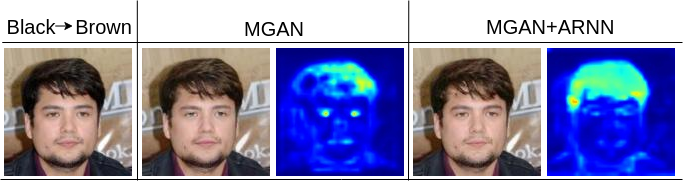}
\includegraphics[width=0.8\linewidth]{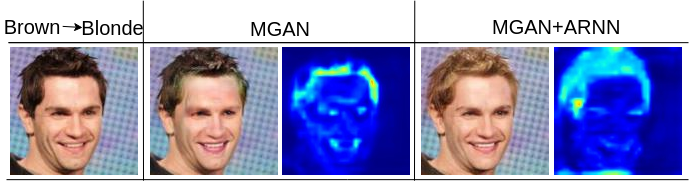}
\includegraphics[width=0.8\linewidth]{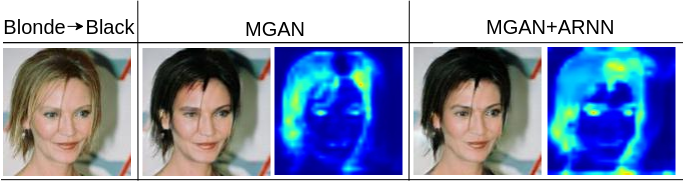}
\includegraphics[width=0.8\linewidth]{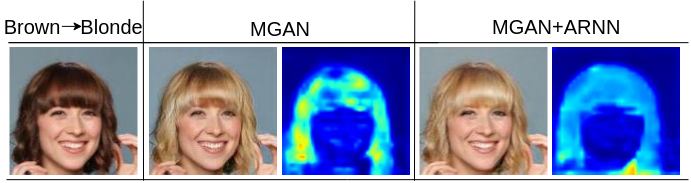}
\caption{{\bf Qualitative Results for Image Generation.} Attention masks generated by MGAN and MGAN+ARNN are shown. Notice that the hair mask is more uniform for MGAN+ARNN as it is able to encode structural dependencies in the attention mask. For detailed explanation see Section 4.4 of the main paper.}
\label{fig:gan1}
\end{figure}

\begin{figure}[H]
\centering
\includegraphics[width=0.8\linewidth]{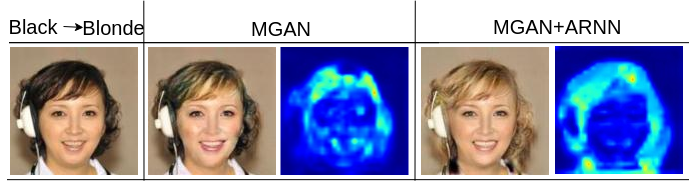}
\includegraphics[width=0.8\linewidth]{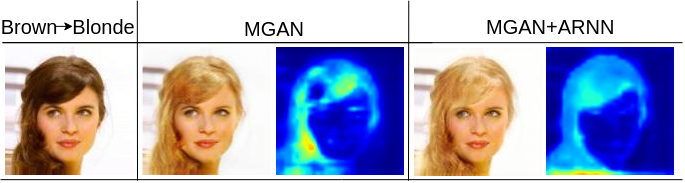}
\includegraphics[width=0.8\linewidth]{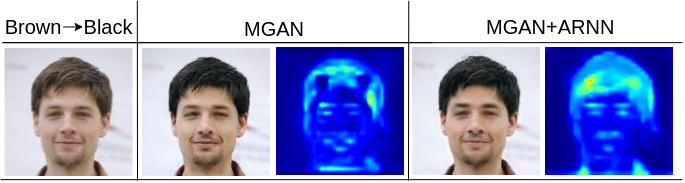}
\includegraphics[width=0.8\linewidth]{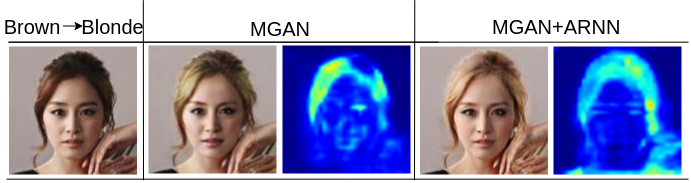}
\caption{{\bf Qualitative Results for Image Generation.} Attention masks generated by MGAN and MGAN+ARNN are shown. Notice that the hair mask is more uniform for MGAN+ARNN as it is able to encode structural dependencies in the attention mask. For detailed explanation see Section 4.4 of the main paper.}
\label{fig:gan2}
\end{figure}

Please refer to Section 4.4 of the main paper for task definition. Figures \ref{fig:gan1} and \ref{fig:gan2} show the attention masks generated by MGAN and MGAN+ARNN for the task of \emph{hair color} transformation. MGAN+ARNN encodes structural dependencies in the attention values, which is evident from the more uniform and continuous attention masks. MGAN, on the other hand, has sharp discontinuities which, in some cases, leads to less accurate hair color transformations.

\twocolumn

{\small
\bibliographystyle{ieee}
\bibliography{egbib}

\begin{thebibliography}{10}\itemsep=-1pt

\bibitem{andreas2016cvpr}
J.~Andreas, M.~Rohrbach, T.~Darrell, and D.~Klein.
\newblock Neural module networks.
\newblock In {\em IEEE Conference on Computer Vision and Pattern Recognition},
  2016.

\bibitem{antol2015iccv}
S.~Antol, A.~Agrawal, J.~Lu, M.~Mitchell, D.~Batra, C.~L. Zitnick, and
  D.~Parikh.
\newblock Vqa: Visual question answering.
\newblock In {\em IEEE International Conference on Computer Vision}, pages
  2425--2433, 2015.

\bibitem{bahdanau2014neural}
D.~Bahdanau, K.~Cho, and Y.~Bengio.
\newblock Neural machine translation by jointly learning to align and
  translate.
\newblock In {\em International Conference on Learning Representations}, 2015.

\bibitem{chen2017sca}
L.~Chen, H.~Zhang, J.~Xiao, L.~Nie, J.~Shao, W.~Liu, and T.-S. Chua.
\newblock Sca-cnn: Spatial and channel-wise attention in convolutional networks
  for image captioning.
\newblock In {\em IEEE Conference on Computer Vision and Pattern Recognition},
  pages 6298--6306. IEEE, 2017.

\bibitem{fukui2016multimodal}
A.~Fukui, D.~H. Park, D.~Yang, A.~Rohrbach, T.~Darrell, and M.~Rohrbach.
\newblock Multimodal compact bilinear pooling for visual question answering and
  visual grounding.
\newblock In {\em Conference on Empirical Methods in Natural Language
  Processing}, pages 457--468. ACL, 2016.

\bibitem{goodfellow2014generative}
I.~Goodfellow, J.~Pouget-Abadie, M.~Mirza, B.~Xu, D.~Warde-Farley, S.~Ozair,
  A.~Courville, and Y.~Bengio.
\newblock Generative adversarial nets.
\newblock In {\em Advances in Neural Information Processing Systems}, pages
  2672--2680, 2014.

\bibitem{goyal2017making}
Y.~Goyal, T.~Khot, D.~Summers-Stay, D.~Batra, and D.~Parikh.
\newblock Making the v in vqa matter: Elevating the role of image understanding
  in visual question answering.
\newblock In {\em IEEE Conference on Computer Vision and Pattern Recognition},
  pages 6325--6334, 2017.

\bibitem{he2016deep}
K.~He, X.~Zhang, S.~Ren, and J.~Sun.
\newblock Deep residual learning for image recognition.
\newblock In {\em IEEE Conference on Computer Vision and Pattern Recognition},
  pages 770--778, 2016.

\bibitem{hochreiter1997long}
S.~Hochreiter and J.~Schmidhuber.
\newblock Long short-term memory.
\newblock {\em Neural computation}, 9(8):1735--1780, 1997.

\bibitem{hu2018eccv}
R.~Hu, J.~Andreas, K.~Saenko, and T.~Darrell.
\newblock Explainable neural computation via stack neural module networks.
\newblock In {\em European Conference on Computer Vision}, 2018.

\bibitem{jaderberg2015spatial}
M.~Jaderberg, K.~Simonyan, A.~Zisserman, et~al.
\newblock Spatial transformer networks.
\newblock In {\em Advances in Neural Information Processing Systems}, pages
  2017--2025, 2015.

\bibitem{johnson2017cvpr}
J.~Johnson, B.~Hariharan, L.~van~der Maaten, F.-F. Li, L.~Zitnick, and
  R.~Girshick.
\newblock Clevr: A diagnostic dataset for compositional language and elementary
  visual reasoning.
\newblock In {\em IEEE Conference on Computer Vision and Pattern Recognition},
  2017.

\bibitem{johnson2016cvpr}
J.~Johnson, A.~Karpathy, and L.~Fei-Fei.
\newblock Densecap: Fully convolutional localization networks for dense
  captioning.
\newblock In {\em IEEE Conference on Computer Vision and Pattern Recognition},
  2016.

\bibitem{kingma2014auto}
D.~P. Kingma and M.~Welling.
\newblock Auto-encoding variational bayes.
\newblock {\em stat}, 1050:10, 2014.

\bibitem{krizhevsky2009learning}
A.~Krizhevsky.
\newblock Learning multiple layers of features from tiny images.
\newblock Technical report, Citeseer, 2009.

\bibitem{krizhevsky2012nips}
A.~Krizhevsky, I.~Sutskever, and G.~E. Hinton.
\newblock Imagenet classification with deep convolutional neural networks.
\newblock In {\em Advances in Neural Information Processing Systems}, pages
  1097--1105, 2012.

\bibitem{lecun1998gradient}
Y.~LeCun, L.~Bottou, Y.~Bengio, and P.~Haffner.
\newblock Gradient-based learning applied to document recognition.
\newblock {\em Proceedings of the IEEE}, 86(11):2278--2324, 1998.

\bibitem{lin2014microsoft}
T.-Y. Lin, M.~Maire, S.~Belongie, J.~Hays, P.~Perona, D.~Ramanan,
  P.~Doll{\'a}r, and C.~L. Zitnick.
\newblock Microsoft coco: Common objects in context.
\newblock In {\em European conference on computer vision}, pages 740--755,
  2014.

\bibitem{liu2017attention}
C.~Liu, J.~Mao, F.~Sha, and A.~L. Yuille.
\newblock Attention correctness in neural image captioning.
\newblock In {\em AAAI}, pages 4176--4182, 2017.

\bibitem{liu2015faceattributes}
Z.~Liu, P.~Luo, X.~Wang, and X.~Tang.
\newblock Deep learning face attributes in the wild.
\newblock In {\em IEEE International Conference on Computer Vision}, 2015.

\bibitem{lu2016hierarchical}
J.~Lu, J.~Yang, D.~Batra, and D.~Parikh.
\newblock Hierarchical question-image co-attention for visual question
  answering.
\newblock In {\em Advances In Neural Information Processing Systems}, pages
  289--297, 2016.

\bibitem{rezende2014stochastic}
D.~J. Rezende, S.~Mohamed, and D.~Wierstra.
\newblock Stochastic backpropagation and approximate inference in deep
  generative models.
\newblock {\em International Conference on Machine Learning}, 2014.

\bibitem{salimans2017pixelcnn++}
T.~Salimans, A.~Karpathy, X.~Chen, and D.~P. Kingma.
\newblock Pixelcnn++: Improving the pixelcnn with discretized logistic mixture
  likelihood and other modifications.
\newblock {\em ICLR}, 2017.

\bibitem{seo2017visual}
P.~H. Seo, A.~Lehrmann, B.~Han, and L.~Sigal.
\newblock Visual reference resolution using attention memory for visual dialog.
\newblock In {\em Advances in Neural Information Processing Systems}, pages
  3719--3729, 2017.

\bibitem{seo2016progressive}
P.~H. Seo, Z.~Lin, S.~Cohen, X.~Shen, and B.~Han.
\newblock Progressive attention networks for visual attribute prediction.
\newblock {\em British Machine Vision Conference}, 2018.

\bibitem{shankar2018posterior}
S.~Shankar and S.~Sarawagi.
\newblock Posterior attention models for sequence to sequence learning.
\newblock In {\em International Conference on Learning Representations}, 2019.

\bibitem{tang2014nips}
Y.~Tang, N.~Srivastava, and R.~R. Salakhutdinov.
\newblock Learning generative models with visual attention.
\newblock In {\em Advances in Neural Information Processing Systems}, 2014.

\bibitem{tommasi2014bmvc}
T.~Tommasi, A.~Mallya, B.~Plummer, S.~Lazebnik, A.~Berg, and T.~Berg.
\newblock Solving visual madlibs with multiple cues.
\newblock In {\em British Machine Vision Conference}, 2016.

\bibitem{van2016conditional}
A.~van~den Oord, N.~Kalchbrenner, L.~Espeholt, O.~Vinyals, A.~Graves, et~al.
\newblock Conditional image generation with pixelcnn decoders.
\newblock In {\em Advances in Neural Information Processing Systems}, pages
  4790--4798, 2016.

\bibitem{van2016pixel}
A.~Van~Oord, N.~Kalchbrenner, and K.~Kavukcuoglu.
\newblock Pixel recurrent neural networks.
\newblock In {\em IEEE International Conference on Machine Learning}, pages
  1747--1756, 2016.

\bibitem{venugopalan2015sequence}
S.~Venugopalan, M.~Rohrbach, J.~Donahue, R.~Mooney, T.~Darrell, and K.~Saenko.
\newblock Sequence to sequence-video to text.
\newblock In {\em Proceedings of the IEEE international conference on computer
  vision}, pages 4534--4542, 2015.

\bibitem{woo2018cbam}
S.~Woo, J.~Park, J.-Y. Lee, and I.~So~Kweon.
\newblock Cbam: Convolutional block attention module.
\newblock In {\em European Conference on Computer Vision}, pages 3--19, 2018.

\bibitem{xiao2016sun}
J.~Xiao, K.~A. Ehinger, J.~Hays, A.~Torralba, and A.~Oliva.
\newblock Sun database: Exploring a large collection of scene categories.
\newblock {\em International Journal of Computer Vision}, 119(1):3--22, 2016.

\bibitem{xu2016ask}
H.~Xu and K.~Saenko.
\newblock Ask, attend and answer: Exploring question-guided spatial attention
  for visual question answering.
\newblock In {\em European Conference on Computer Vision}, pages 451--466,
  2016.

\bibitem{xu2015show}
K.~Xu, J.~Ba, R.~Kiros, K.~Cho, A.~Courville, R.~Salakhudinov, R.~Zemel, and
  Y.~Bengio.
\newblock Show, attend and tell: Neural image caption generation with visual
  attention.
\newblock In {\em International Conference on Machine Learning}, pages
  2048--2057, 2015.

\bibitem{yang2016stacked}
Z.~Yang, X.~He, J.~Gao, L.~Deng, and A.~Smola.
\newblock Stacked attention networks for image question answering.
\newblock In {\em IEEE Conference on Computer Vision and Pattern Recognition},
  pages 21--29, 2016.

\bibitem{you2016image}
Q.~You, H.~Jin, Z.~Wang, C.~Fang, and J.~Luo.
\newblock Image captioning with semantic attention.
\newblock In {\em IEEE conference on computer vision and pattern recognition},
  pages 4651--4659, 2016.

\bibitem{zhang2018arxiv}
H.~Zhang, I.~Goodfellow, D.~Metaxas, and A.~Odena.
\newblock Self-attention generative adversarial networks.
\newblock In {\em arXiv preprint arXiv:1805.08318}, 2018.

\bibitem{zhang2017stackgan}
H.~Zhang, T.~Xu, H.~Li, S.~Zhang, X.~Huang, X.~Wang, and D.~Metaxas.
\newblock Stackgan: Text to photo-realistic image synthesis with stacked
  generative adversarial networks.
\newblock {\em IEEE International Conference on Computer Vision}, 2017.

\bibitem{zhao2018modular}
B.~Zhao, B.~Chang, Z.~Jie, and L.~Sigal.
\newblock Modular generative adversarial networks.
\newblock {\em European Conference on Computer Vision}, 2018.

\bibitem{zheng2017learning}
H.~Zheng, J.~Fu, T.~Mei, and J.~Luo.
\newblock Learning multi-attention convolutional neural network for
  fine-grained image recognition.
\newblock In {\em IEEE International Conference on Computer Vision}, volume~6,
  2017.

\bibitem{zhu2016visual7w}
Y.~Zhu, O.~Groth, M.~Bernstein, and L.~Fei-Fei.
\newblock Visual7w: Grounded question answering in images.
\newblock In {\em IEEE Conference on Computer Vision and Pattern Recognition},
  pages 4995--5004, 2016.

\end{thebibliography}
}

\end{document}